\documentclass{article}

\PassOptionsToPackage{numbers, compress}{natbib}

\usepackage[preprint]{neurips_2026}

\usepackage[utf8]{inputenc} 
\usepackage[T1]{fontenc}    
\usepackage{hyperref}       
\usepackage{url}            
\usepackage{booktabs}       
\usepackage{amsfonts}       
\usepackage{nicefrac}       
\usepackage{microtype}  
\usepackage{enumitem}
\usepackage{xcolor}         
\usepackage{graphicx}       
\usepackage{amsmath}
\usepackage{pifont}
\usepackage{subcaption}
\newcommand{\cmark}{\ding{51}}%
\newcommand{\xmark}{\ding{55}}%
\usepackage{booktabs}
\usepackage{tabularx}
\usepackage{array}
\usepackage{makecell}
\usepackage{wrapfig}
\usepackage{makecell}
\usepackage{multirow}
\usepackage{booktabs}
\usepackage{adjustbox}
\usepackage{graphicx}
\usepackage{float}
\usepackage[table]{xcolor}
\newcolumntype{Y}{>{\raggedright\arraybackslash}X}
\newcolumntype{L}[1]{>{\raggedright\arraybackslash}p{#1}}

\usepackage[most]{tcolorbox}
\usepackage{xcolor}
\usepackage{listings}
\usepackage{enumitem}
\definecolor{promptbg}{HTML}{FAFAFA}
\definecolor{promptborder}{HTML}{D9D9D9}
\definecolor{prompttitlebg}{HTML}{F0F0F0}

\newtcblisting{promptbox}[2][]{
  enhanced,
  breakable,
  colback=promptbg,
  colframe=promptborder,
  colbacktitle=prompttitlebg,
  title={#2},
  fonttitle=\bfseries\footnotesize,
  coltitle=black,
  listing only,
  listing options={
    basicstyle=\ttfamily\scriptsize,
    breaklines=true,
    columns=fullflexible,
    keepspaces=true,
    showstringspaces=false,
    tabsize=2,
    xleftmargin=0.5mm
  },
  left=1.5mm,
  right=1.5mm,
  top=1.2mm,
  bottom=1.2mm,
  boxrule=0.35pt,
  arc=1mm,
  before skip=6pt,
  after skip=6pt,
  #1
}

\title{Know You Before You Speak: User-State Modeling for LLM Personalization in Multi-Turn Conversation}

\author{%
  \textbf{Jiani Luo}$^{1}$,
  \textbf{Xiaoyan Zhao}$^{1}$\thanks{Corresponding authors.}~,
  \textbf{Yang Zhang}$^{1}$\footnotemark[1]~,
  \textbf{Shuyi Miao}$^{2}$ \\
  \textbf{Bingbing Xu}$^{3}$,
  \textbf{Stefan Konigorski}$^{4}$,
  \textbf{Tat-Seng Chua}$^{1}$ \\
  \\
  $^{1}$School of Computing, National University of Singapore, Singapore \\
  $^{2}$School of Artificial Intelligence, Beihang University, Beijing, China \\
  $^{3}$Institute of Computing Technology, Chinese Academy of Sciences, Beijing, China \\
  $^{4}$German Institute of Human Nutrition Potsdam-Rehbruecke, Germany \\
  \texttt{xzhao@se.cuhk.edu.hk, zhangy@nus.edu.sg, e1616036@u.nus.edu}
}

\begin{document}

\maketitle

\begin{abstract}
Personalized dialogue requires more than recalling explicit user histories: systems also need to infer hidden user states that evolve through interaction and shape appropriate response strategies. Existing memory- and profile-based methods primarily reuse observable user information, offering limited support for modeling user-state dynamics or selecting actions based on how they shape future user states. We propose PUMA (Prospective User-state Modeling for Action selection), a framework grounded in the Free Energy Principle (FEP) that formulates personalization as decision-making under partial observability, centered on an explicit user state model that captures latent user states and their action-conditioned dynamics. At each turn, PUMA maintains a belief over the user's hidden state, refines the user state model for observation generation and action-conditioned state transition, and selects dialogue actions by minimizing expected free energy—balancing epistemic and pragmatic objectives under a unified criterion. This formulation shifts personalization from passive memory retrieval to model-based decision-making over user evolution. We instantiate PUMA on healthcare-oriented counseling and motivational interviewing benchmarks with latent state annotations for rigorous evaluation. 
Experiments show that PUMA improves long-horizon dialogue outcomes while maintaining strong response quality, and a cross-dataset study demonstrates more reliable user-state estimation and next-state prediction. 
Our code is available at: \url{https://github.com/Annie1161/PUMA}. 


\end{abstract}

\section{Introduction}
\label{sec:intro}
%


Personalized dialogue systems aim to generate responses that remain coherent, adaptive, and user-specific over long-term interactions \citep{zhao2025reinforced, li2025hello, chen2017survey}. 
They are important in applications such as personal assistants, education, healthcare communication, mental health support, and long-term decision support, where effective interaction depends on understanding both the current query and the user's evolving context. 
Recent LLM-based systems \citep{zhao2026nextquill,qiu2025latent,zhao2025steerx,wang2026think,zhang2026nextmem,yang2025reliable} have advanced personalization by incorporating long-term memory, user profiles, and retrieved interaction histories, enabling models to reuse explicit user information such as preferences, decisions, and prior dialogue content.

However, long-term personalization cannot be reduced to recalling explicitly stated user information. The same utterance may require fundamentally different responses depending on the user's latent state~\cite{cho2022implicit}. Consider a user who says, "I've been taking the new medication for two weeks". The appropriate response depends on whether the user is anxious about side effects, confidently tracking progress, or subtly reconsidering discontinuation. These differences reflect latent factors such as uncertainty, engagement, and readiness to act~\citep{kodama2021internal}, which are only partially expressed in language but critically shape how an utterance should be interpreted and how the system should respond \citep{cho2022implicit, tang2023contrastive, lu2023miracle}. Therefore, effective personalization requires not only tracking what the user has said, but also inferring hidden user states and reasoning about how system actions influence their evolution over time. This points to the need for a user state model—a predictive representation of the user as a latent dynamical system whose state evolves under system interventions—as the foundation for long-term personalization.

Existing personalized dialogue systems remain limited from this perspective. Memory-augmented and profile-based methods \citep{zhao2025exploring,huang2026mem,yu2026memweaver, xu2024crafting} operate over explicit textual records, making them effective for recalling stable facts but insufficient for modeling latent state evolution. While some approaches introduce latent variables, these representations are typically static or descriptive, failing to capture the continuous and action-dependent dynamics of user behavior \citep{wu2026humanlm, qiu2025latent, liu2025persona}. As a result, existing systems implicitly treat personalization as a retrieval or conditioning problem, rather than a sequential decision-making problem over a partially observable user. In short, current methods either lack a user state model entirely or rely on impoverished, non-predictive surrogates—leaving prospective reasoning about user evolution under system actions out of reach.

The Free Energy Principle (FEP) and active inference \citep{friston2006free, friston2016active} provide a principled decision-theoretic foundation for addressing this gap. Under FEP, an agent operates in a partially observable environment by maintaining and updating beliefs over hidden states, and selecting actions by minimizing expected free energy (EFE), which balances uncertainty reduction (epistemic value) and goal-directed behavior (pragmatic value). Since EFE is evaluated over action-conditioned dynamics, actions are chosen based on both their immediate effects and their influence on future state beliefs. This formulation naturally aligns with personalized dialogue, where the user is a latent dynamical system and system responses act as interventions that shape its evolution.

Building on this perspective, we propose \textbf{PUMA} (\textbf{P}rospective \textbf{U}ser-state \textbf{M}odeling for \textbf{A}ction selection), an FEP-grounded framework for personalized dialogue. At the core of PUMA lies a user state model, defined as a predictive model over latent user states and their action-conditioned dynamics, capturing how the user evolves in response to system interventions. This goes beyond static state estimation by explicitly modeling state transitions induced by dialogue actions.
At each dialogue turn, PUMA maintains a belief over the user's latent state and a world model of user-state dynamics, updates them based on observations, and evaluates candidate responses by estimating their expected impact on future states through an approximation of expected free energy. This enables joint inference and control within a unified decision framework grounded in the FEP.
This formulation shifts personalization from passive memory retrieval to model-based decision-making over user evolution. Rather than selecting responses based on retrieved context or static preferences, PUMA evaluates candidate actions by their expected influence on future user trajectories, balancing epistemic and pragmatic objectives under a single principled criterion.

We instantiate PUMA on healthcare dialogue benchmarks, which provide the latent user state annotations necessary for evaluating belief tracking and long-horizon decision quality. Experimental results show that explicitly modeling user state dynamics and performing FEP-based action evaluation significantly improve both user state tracking and long-horizon dialogue performance compared with strong LLM-based baselines.

Our main contributions are summarized as follows:
\begin{itemize}[itemsep=1pt, topsep=1pt, parsep=0pt, partopsep=0pt,leftmargin=*]
\item We introduce a user state model for personalized dialogue that, unlike static persona, memory-based or profile-based   representations in prior work, captures latent user states and their action-conditioned evolution over time.

\item We formulate personalized dialogue under the Free Energy Principle and develop PUMA, an FEP-grounded framework that unifies belief updating and action selection via expected free energy, enabling prospective decision-making by reasoning over future user states.

\item We empirically evaluate PUMA on healthcare dialogue benchmarks, where FEP-grounded user state modeling and response selection yield consistent gains in both belief tracking and long-horizon dialogue outcomes over strong LLM-based baselines.

\end{itemize}
\section{Related Work}
\label{sec:related_work}

\subsection{Personalized Dialogue Systems}
\label{sec:rw_personalized_dialogue}

Personalized dialogue systems aim to generate responses that remain coherent, adaptive, and user-specific across long-term interactions.
Existing studies can be broadly categorized into profile-based personalization, memory-augmented dialogue systems, long-context personalization, and latent-variable dialogue modeling.
Profile-, memory-, and long-context-based methods represent users with persona descriptions, preference profiles, structured attributes~\citep{zhao2026nextquill,shen2024pmg,zhao2025reinforced,wu2025rlpf,miao2026unidetect}, short- or long-term memories~\citep{zhang2026memskill,huang2026mem,yu2026memweaver,zhang2025prime,zhong2024memorybank}, or extended interaction histories~\citep{shi2025answering,cai2025large,lee2024aligning,wang2025msl,sarthi2024raptor,zhang2026reinforced, shao2024average}, and inject them through conditioning, retrieval, or long-context prompting to reuse past interactions, stated preferences, decisions, and personal facts.
In parallel, latent-variable and state-aware dialogue models introduce hidden representations to capture implicit factors such as user intent, emotion, preference, dialogue state, or conversational goals~\citep{zhao2026nextquill,zhang2026explicit,wu2026humanlm,qiu2025measuring,qiu2025latent}.
These methods move beyond purely explicit user records by recognizing that observable utterances may only partially reveal the user's underlying condition.
Existing methods mainly reuse user records or context, with limited support for action-conditioned latent state dynamics.
PUMA explicitly models latent user state separately from memory, enabling active state-aware action selection beyond passive history reuse.

\subsection{Free Energy Principle and Its Applications}
\label{sec:rw_fep}

The Free Energy Principle provides a principled account of adaptive behavior under uncertainty, 
viewing perception and action as a unified inference process over hidden states of the world 
\citep{friston2010free,friston2010action,friston2017active}.
Under this view, an agent updates its beliefs from observations and acts to make future observations 
more consistent with its preferences.
Active inference operationalizes this process by selecting actions or policies according to expected 
free energy, which integrates goal-directed value with uncertainty reduction 
\citep{friston2015active,friston2016active}.
This perspective has been extended to robotics, reinforcement learning, control, and AI agents, 
emphasizing belief updating and uncertainty-aware planning under partial observability 
\citep{ma2026odar,he2024large,pezzulo2024generating,fang2023large}.
Recent work has further connected active inference with human-facing LLM and interactive systems, 
including reliable medical prompting \citep{shusterman2025active}, active preference learning and 
inference \citep{mahmud2025maple,piriyakulkij2023active,liu2026mtp}, and human--computer interaction modeling 
\citep{murray2025active}.
Existing FEP-inspired systems mainly apply active inference to external environments, task states, or relatively stable human-related variables, rarely treating the evolving user state itself as the modeling target.
In contrast, PUMA applies FEP to personalized dialogue by making temporally continuous and action-sensitive user state the core object of active-inference modeling.

\section{Preliminaries}
\label{sec:preliminaries}

The Free Energy Principle (FEP), originating from theoretical neuroscience and cognitive science, explains how biological agents adapt through perception, learning, and action when interacting with their external environment~\citep{friston2006free, friston2015active}.
This principle is commonly operationalized through two related objectives: variational free energy \(F_t\) (Section~\ref{sec:fep}) for retrospective belief and internal world model updating over the present state, and expected free energy \(G_t\) (Section~\ref{sec:active_inf}) for prospective action selection over future outcomes. We briefly introduce these two objectives below.
\subsection{Variational Free Energy for Belief and World-Model Updating}
\label{sec:fep}

FEP views an agent as interacting with a partially observed environment, where the hidden state \(s_t\) generates the observation \(o_t\) but cannot be directly accessed.
The agent therefore maintains two internal quantities: a belief over the hidden state, representing its current uncertain understanding of the hidden state, and a world model of how the environment works.
After receiving an observation, the agent updates its belief over the hidden state and refines its internal world model to better explain the observation.
Variational free energy \(F_t\) provides a scalar objective for this process.
In general, it evaluates how well the agent's approximate belief \(q(s_t)\) and world model \(p(o_t, s_t)\) explain observations under uncertainty:
\begin{equation}
F_t
=
\mathbb{E}_{q(s_t)}
\left[
\log q(s_t) - \log p(o_t, s_t)
\right].
\label{eq:vfe_general}
\end{equation}

Minimizing \(F_t\) supports belief updating and world model learning, enabling more plausible belief over user state and a more accurate account of the environment's underlying dynamics.

\subsection{Expected Free Energy for Action Selection} \label{sec:active_inf} 
While variational free energy concerns updating beliefs and models after observations are received, action selection requires reasoning about outcomes that have not yet occurred.
Expected free energy \(G_t(\pi)\) extends the free-energy objective to this prospective setting by evaluating the anticipated consequences of candidate policies under the agent's belief and world model.

For each candidate policy \(\pi\), the agent predicts possible future states and observations, and estimates whether the policy is expected to reduce uncertainty and lead to goal-directed outcomes.
A policy with lower expected free energy is therefore expected to be both informative and goal-directed:
\begin{equation}
\pi^\star = \arg\min_\pi G_t(\pi).
\label{eq:efe_policy}
\end{equation}
Through this, expected free energy provides a general objective for action selection under uncertainty.

\begin{figure}[t]
\vspace{-2em}
    \centering
    \includegraphics[width=1\linewidth]{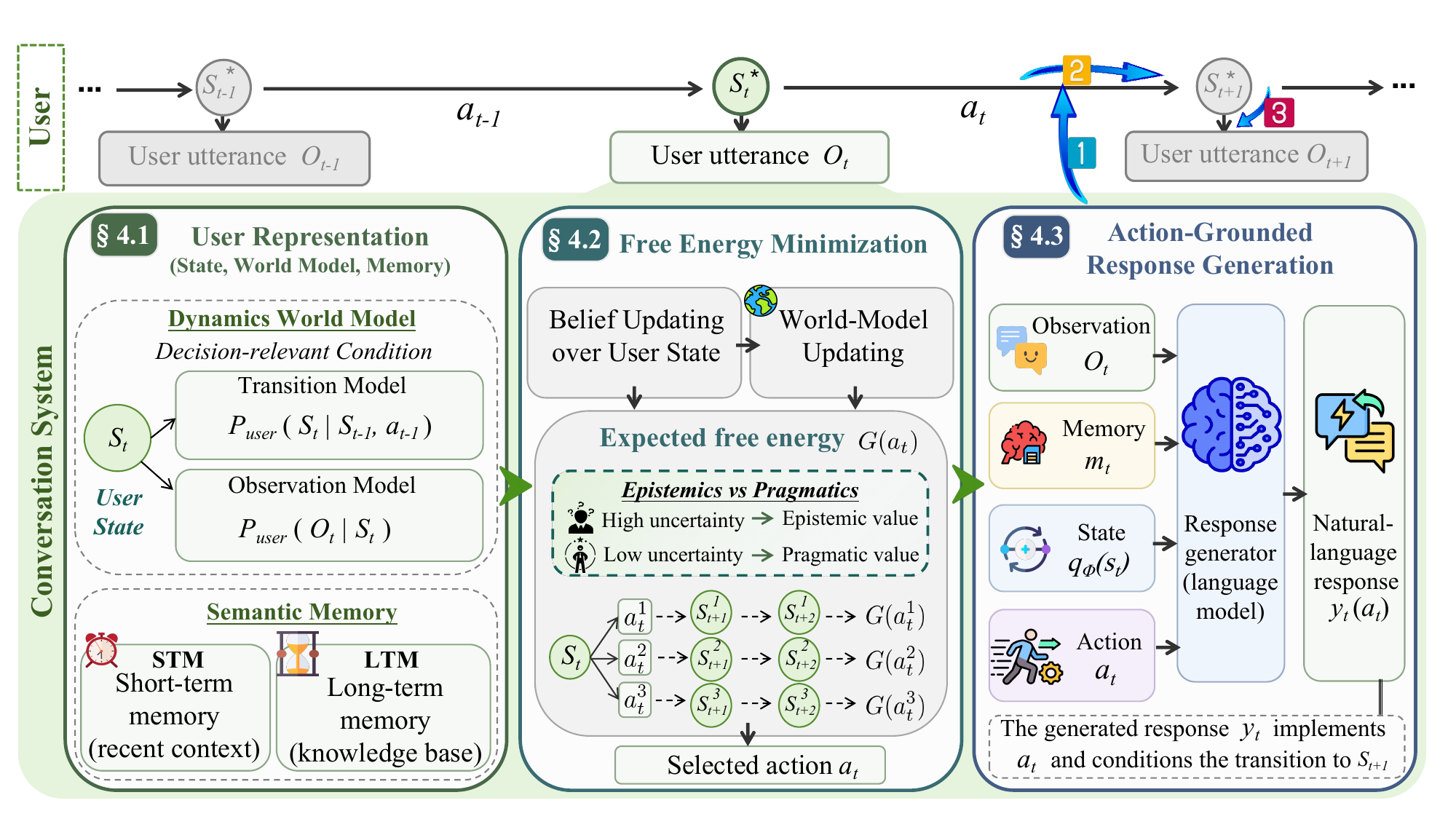}
    \caption{Overview of PUMA. Given a user utterance, the system updates its belief over the hidden user state and maintains a world model of user-state dynamics.
It then uses this world model to evaluate candidate actions with expected free energy, selecting actions that reduce uncertainty and guide goal-directed outcomes.
Finally, the system produces a natural-language response, which gives rise to a user-state transition and the next user utterance, thereby starting the next interaction cycle.
}
    \label{fig:framework}
\vspace{-1em}
\end{figure}
\section{Method}
\label{sec:method}

Personalization is not merely about maintaining observable continuity in dialogue content; it also requires accounting for unobservable user-specific factors that influence user interpretation and response. 
We therefore propose \textbf{PUMA}, which models the user as a partially observable dynamical system, maintains a belief over user state, tracks the user's evolving state through a world model, and selects actions that balance user uncertainty reduction and system goal-directed behavior. 
We first introduce our overall framework in Section~\ref{sec:framework}, and then detail user representation, belief refinement, world-model updating, action selection, and response generation in Sections~\ref{sec:latent_user_modeling}--\ref{sec:response_generation}.



\subsection{Overview: Framework of PUMA}
\label{sec:framework}

In personalized dialogue, some important user-specific factors are implicit in language, making personalization a decision-making problem under partial observability.
PUMA addresses this problem by maintaining a belief over the user's evolving state, refining a world model through interaction, and selecting actions according to their expected consequences.
Figure~\ref{fig:framework} illustrates the overall framework.

\textbf{User Representation.}
Section~\ref{sec:latent_user_modeling} defines the user representation by separating user state, world model, and semantic memory, which together capture the user's hidden condition, action-conditioned dynamics, and content-level evidence for personalization.

\textbf{Free Energy Minimization.}
Building on this representation, Section~\ref{sec:inference_learning_planning} describes how the system updates its belief over the user's state, refines the world model of user-state dynamics, and selects actions by balancing uncertainty reduction with goal-directed dialogue behavior.

\textbf{Action-Grounded Response Generation.}
Given the selected action and updated belief over the user state, Section~\ref{sec:response_generation} explains how the system generates a natural-language response grounded in the current observation, retrieved memory, and future-oriented dialogue action. 



\subsection{User Representation}
\label{sec:latent_user_modeling}

This section defines the core user representation in our framework.
We distinguish three related but different concepts: the user state, the world model, and semantic memory.
The user state captures the user's current internal condition, the world model describes how this state evolves and generates observations, and memory provides concrete personalized content for response generation.

\paragraph{User State}
\label{sec:latent_belief}
Personalized dialogue requires modeling not only content-level user information but also the user's evolving internal condition, which may shape utterances without being explicitly stated.
We assume that at each turn \(t\), the user has a true \(s_t^\ast\); this state represents the user's current decision-relevant condition.
Since \(s_t^\ast\) is not directly accessible to the system, our system maintains an internal representation \(s_t\), inferred from the dialogue history and the current observation.
The user state \(s_t\) evolves over time and is action-relevant, enabling the system to interpret the current observation \(o_t\) beyond its explicit content and allowing system actions to shape its future trajectory.

\paragraph{World Model}
\label{sec:method_world_model}

Given the user state, we introduce a world model to describe how user states evolve under system actions and how they are expressed in language.
Formally, we factorize the world model as
\begin{equation}
p_\theta(o_t, s_t \mid s_{t-1}, a_{t-1})
=
p_\theta(s_t \mid s_{t-1}, a_{t-1})
p_\theta(o_t \mid s_t).
\label{eq:world_model}
\end{equation}
This factorization separates two roles: \(p_\theta(s_t \mid s_{t-1}, a_{t-1})\) captures action-conditioned state evolution, while \(p_\theta(o_t \mid s_t)\) captures how the user state generates the observed utterance.

\noindent\emph{Transition model.}
The model \(p_\theta(s_t \mid s_{t-1}, a_{t-1})\) captures how the user's state evolves across turns under system actions.
By modeling the effect of action \(a_{t-1}\) on the next state, it provides both the predictive component for future-state forecasting and the control target for action selection.
\noindent\emph{Observation model.}
The observation model \(p_\theta(o_t \mid s_t)\) captures how a user state is expressed as an observable utterance.
It links the hidden state to the actual response \(o_t\), providing the inferential component for updating the system's belief over \(s_t\).

Together, these two components define the dynamic structure of user-state personalization and support the action selection described in Section~\ref{sec:inference_learning_planning}.

\paragraph{Semantic Memory}
\label{sec:method_memory}

The user state \(s_t\) captures the user's evolving, action-relevant condition, but does not store concrete semantic content.
We therefore maintain a two-level semantic memory \(M_t=\{M_t^{\mathrm{STM}},M_t^{\mathrm{LTM}}\}\), where short-term memory stores session-local context and long-term memory stores cross-session preferences, histories, goals, and behavioral regularities.

At each turn, the system retrieves relevant memory conditioned on the current belief, selected action, and dialogue context:
\begin{equation}
m_t = \mathrm{Retrieve}(M_t^{\mathrm{STM}}, M_t^{\mathrm{LTM}} \mid q_\phi(s_t), a_t, o_{\le t}).
\end{equation}
The retrieved memory \(m_t\) grounds response generation with content-level personalization while keeping user-state tracking separate from semantic storage.
\subsection{Free Energy Minimization}
\label{sec:inference_learning_planning}

Based on the user representation above, the system performs a Free Energy Minimization over turns.
It first updates the belief over the user state from the new observation, then refines the world model for user-state dynamics, and finally selects actions through the evaluation of future user-state trajectories.

\paragraph{Belief Updating over User State}
\label{sec:method_inference}


Because the true user state \(s_t^\ast\) is unobserved, the system introduces an internal user-state variable \(s_t\) to approximate it, and maintains a variational belief \(q_\phi(s_t)\) inferred from the history and the current utterance \(o_t\).
Instantiating the general variational-free-energy objective in Eq.~\ref{eq:vfe_general} with the world model in Eq.~\ref{eq:world_model}, the turn-level belief update is defined as
\begin{equation}
q_\phi(s_t)
=
\arg\min_q
\left[
D_{\mathrm{KL}}\big(q(s_t)\,\|\,\tilde p_\theta(s_t)\big)
-
\mathbb{E}_{q(s_t)}
\big[\log p_\theta(o_t \mid s_t)\big]
\right],
\label{eq:posterior_inference}
\end{equation}
where \(\tilde p_\theta(s_t)\) denotes the predictive prior over the current user state induced by the previous belief and the previous system action.
The derivation of Eq.~\ref{eq:posterior_inference}, as well as the justification for introducing \(q_\phi(s_t)\) under the world model, is provided in Appendix~\ref{app:posterior_derivation}.
The KL term preserves consistency with predicted state dynamics, while the likelihood term grounds the update in the observed utterance, enabling belief updating to balance dynamic plausibility with explanatory accuracy.

\paragraph{World-Model Updating}
\label{sec:world_model_learning}

After observing \(o_t\) and updating \(q_\phi(s_t)\), the system refines the world model using the discrepancy between predicted and observed user responses.
The world-model refinement objective can be obtained by optimizing the observation generation and the action-conditioned state transition.
The derivation from the formulation in Eq.~\ref{eq:posterior_inference} is provided in Appendix~\ref{app:wm_derivation}.
Specifically, we define
\begin{equation}
\mathcal{L}_{\mathrm{WM}}
=
\sum_{t=1}^{T}
\mathbb{E}_{q_\phi(s_t)}
\left[
-\log p_\theta(o_t \mid s_t)
\right]
+
\lambda
\sum_{t=1}^{T-1}
\mathbb{E}_{q_\phi(s_t)}
\left[
-\log p_\theta(s_{t+1} \mid s_t, a_t)
\right],
\label{eq:wm_learning}
\end{equation}
where the first term trains the observation generation model, the second trains the action-conditioned transition model, and \(\lambda\) balances the two.
This objective continuously refines how the world model links user state, user utterance, and system action, enabling more accurate prediction of user dynamics.

\paragraph{Action Selection through Expected Free Energy}
\label{sec:action_selection}

Given the current belief \(q_\phi(s_t)\) and the learned world model, action selection can be seen as a model-based computation: comparing candidate actions by their predicted consequences and choosing the one that best achieves uncertainty reduction and goal-directed behavior.
Following the expected-free-energy principle introduced in Section~\ref{sec:active_inf}, we instantiate \(G(a)\) using the learned world model for prospective action selection, with the detailed derivation provided in Appendix~\ref{app:efe_derivation}. 
\begin{equation}
G(a)
=
\underbrace{\mathbb{E}_{q(o_{t+1}\mid a)}
\left[
H\!\left(q(s_{t+1}\mid o_{t+1},a)\right)
\right]}_{\text{epistemic value}}
+
\underbrace{\mathbb{E}_{q(s_{t+1},o_{t+1}\mid a)}
\left[-\log p_{\mathrm{pref}}(o_{t+1})\right]}_{\text{pragmatic value}},
\end{equation}

The system then selects the action that minimizes expected free energy:
\(a=\arg\min_{a\in\mathcal{A}}G(a)\). Here, \(H(\cdot)\) denotes entropy, measuring the remaining uncertainty about the user's next state after action \(a\) and the predicted user response.
\noindent\emph{Epistemic value} favors actions that are expected to reduce uncertainty about the user.
\noindent\emph{Pragmatic value} favors actions whose predicted consequences are more aligned with the system's preferred outcomes, corresponding to goal-directed value.
Thus, action selection is dynamically calibrated by the system's familiarity with the user, providing an adaptive control mechanism that other personalization methods do not possess.

\subsection{Action-Grounded Response Generation}
\label{sec:response_generation}
The selected action \(a_t\) specifies the dialogue-level behavior, which is then realized as a natural-language response grounded in the belief and personalized content.
The response is generated conditioned on the selected action, the belief over the user state, retrieved memory, and the current user utterance:
\begin{equation}
y_t \sim p_\psi(y_t \mid a_t, q_\phi(s_t), m_t, o_{t}).
\end{equation}
Here, \(a_t\) provides explicit control, \(q_\phi(s_t)\) represents the system's belief over the current user state, and \(m_t\) supplies content-level personalization for grounding the response.
The generated response is delivered to the user, influencing the user's state and initiating the next interaction turn.



\section{Experiments}
\label{sec:experiments}

In this section, we conduct experiments to answer the following research questions (RQs): 
\textbf{RQ1}: Can PUMA select actions that better uncover user needs, and steer the dialogue toward desired future outcomes? 
\textbf{RQ2}: How does each component, including belief, the world model, and EFE-based action selection, contribute to the overall performance? 
\textbf{RQ3}: How well does the proposed user-state modeling mechanism support current-state estimation and next-state prediction across dialogue turns?
\textbf{RQ4}: Can the proposed framework generalize its user-state estimation and transition prediction ability to a different dataset?

\subsection{Experimental Setup}
\label{sec:experimental_setup}
\paragraph{Dataset.}We evaluate on CAMI (Counseling and Motivational Interviewing)~\citep{yang2025cami}, which contains 38 client profiles with complete MI counseling sessions.
Each session is annotated with per-turn stage labels following the Transtheoretical Model \citep{prochaska2008initial}: \emph{precontemplation}, \emph{contemplation}, and \emph{preparation}.
Each profile includes a warmup session for persona construction and subsequent sessions for evaluation.
For dynamic evaluation, we use the precontemplation-start profiles after removing duplicated profiles from the original set.
For offline state-inference evaluation, we use the full profile set, yielding 427 evaluation turns after warmup.

\paragraph{Simulator-based dynamic evaluation.}
Since fixed offline transcripts cannot reveal how clients respond to different counselor policies, we conduct dynamic evaluation with \textit{DynPatient}, a profile-grounded client simulator built from CAMI profiles and annotated sessions.
Given the client profile, current stage, dialogue history, and counselor response, the simulator updates the client state and generates the next utterance.
Details and validation are provided in Appendix~\ref{app:simulator_details}.

\paragraph{Compared Methods.}
We compare PUMA against representative counselor agent baselines.
\textit{DIIR}~\citep{xie-etal-2024-shot-dialogue} uses information integration and reflection.
\textit{CoS}~\citep{sun-etal-2025-rethinking} is a chain-of-suggestion counselor.
\textit{CAMI-base}, \textit{CAMI-text} and \textit{CAMI-full}~\citep{yang2025cami} are the basic, text-only and full CAMI counselors.

\paragraph{Evaluation Metrics.}
We evaluate the framework from two perspectives: dynamic counseling effectiveness and counselor-side response quality.
For dynamic counseling effectiveness, we report \textit{Stage Lift (Lift)}, \textit{Prep Rate (Prep)}, \textit{Trigger Coverage (TrigCov)}, and \textit{Avg Turns (Turns)}.
For counselor-side quality, we use the \textit{MITI Global Scores}~\citep{yang2025cami}, including \textit{Cultivating Change Talk}, \textit{Softening Sustain Talk}, \textit{Partnership}, and \textit{Empathy}, and average them as \textit{MITI-Avg}.
Detailed metric definitions are provided in Appendix~\ref{app:metrics}.

\paragraph{Implementation Details.}
All experiments use \textit{Qwen3-8B}~\citep{yang2025qwen3} and \textit{Llama-3.1-8B}~\citep{grattafiori2024llama} as backbone LLMs via local HuggingFace inference.
For dynamic evaluation, all counselor methods interact with the same fixed client simulator, DynPatient.
To make the system's representation of the user state directly evaluable in practice, we instantiate the user-state space with gold-labeled state annotations from existing datasets.
All results use deterministic greedy decoding with fixed random seeds.
In implementation, PUMA is realized through a set of fixed LLM-based inference modules that instantiate the framework's core components.
The LLM judge for simulator validation and counselor-side quality evaluation uses Qwen3-32B in 4-bit quantization.
The complete PUMA prompt templates are provided in Appendix~\ref{app:prompts_fep}.
The prompt templates for the long-prompt state-inference baseline in RQ3 are provided in Appendix~\ref{app:prompts_baselines_state}.
Additional hyperparameter details are provided in Appendix~\ref{app:hyperparams}, and simulator details are provided in Appendix~\ref{app:simulator_details}.

\subsection{Main Results (RQ1)}
\label{sec:main_results}

In this section, we evaluate whether PUMA can improve dynamic MI-coaching effectiveness.
We focus on whether the model can select appropriate counseling actions, and guide the client toward goal-directed states.
We report the main dynamic counseling results in Table~\ref{tab:dynamic_main}, followed by counselor-side quality evaluation in Table~\ref{tab:miti_global}.
All results are obtained under the same CAMI evaluation protocol.

\paragraph{Dynamic counseling effectiveness.}

Table~\ref{tab:dynamic_main} shows that PUMA achieves the best dynamic counseling performance.
Compared with DIIR, it improves Lift, Prep, and TrigCov while using fewer Turns.
This suggests that state-aware action selection enables more efficient progression toward the desired user state.

\paragraph{Counselor-side MI quality.}

Table~\ref{tab:miti_global} shows that PUMA receives the highest LLM-judge MITI scores among automated methods and remains comparable to the human-reference responses under this evaluation protocol. 
Overall, PUMA improves motivational outcomes while preserving a collaborative and empathetic counseling style.

\begin{table}[h]
\centering
\caption{Dynamic evaluation on CAMI under Qwen3-8B and Llama-3.1-8B backbones. Higher values are better for Lift, Prep, and TrigCov, while lower values are better for Turns.}
\label{tab:dynamic_main}
\scriptsize
\makebox[\linewidth][c]{
\resizebox{1\linewidth}{!}{
\begin{tabular}{@{}lcccccccc@{}}
\toprule
\multirow{2}{*}{\textbf{Counselor}}
& \multicolumn{4}{c}{\textbf{Qwen3-8B}}
& \multicolumn{4}{c}{\textbf{Llama-3.1-8B}} \\
\cmidrule(lr){2-5} \cmidrule(lr){6-9}
& \textbf{Lift}$\uparrow$
& \textbf{Prep}$\uparrow$
& \textbf{TrigCov}$\uparrow$
& \textbf{Turns}$\downarrow$
& \textbf{Lift}$\uparrow$
& \textbf{Prep}$\uparrow$
& \textbf{TrigCov}$\uparrow$
& \textbf{Turns}$\downarrow$ \\
\midrule
CoS~\citep{sun-etal-2025-rethinking}
& 0.17 & 7.0\% & 5.0\% & 19.8
& 0.07 & 3.0\% & 3.0\% & 19.7 \\
DIIR~\citep{xie-etal-2024-shot-dialogue}
& 1.07 & 52.0\% & 34.0\% & 13.1
& 1.17 & 59.0\% & 40.0\% & 11.9 \\
CAMI-base~\citep{yang2025cami}
& 0.62 & 28.0\% & 22.0\% & 16.8
& 0.38 & 17.0\% & 14.2\% & 18.3 \\
CAMI-text~\citep{yang2025cami}
& 0.76 & 31.0\% & 27.0\% & 16.8
& 1.10 & 52.0\% & 34.9\% & 13.9 \\
CAMI-full~\citep{yang2025cami}
& 0.62 & 28.0\% & 23.0\% & 17.0
& 0.72 & 34.0\% & 24.5\% & 15.8 \\
\midrule
\textbf{PUMA}
& \textbf{1.62} & \textbf{75.9\%} & \textbf{62.4\%} & \textbf{12.2}
& \textbf{1.76} & \textbf{83.0\%} & \textbf{63.2\%} & \textbf{10.7} \\
\bottomrule
\end{tabular}
}
}
\vspace{-8pt}
\vspace{-1em}
\end{table}

\begin{table}[h]
\centering
\caption{Counselor-side quality evaluation using MITI Global Scores on CAMI. We report four MI dimensions and their average score; higher scores indicate better counseling quality.}
\label{tab:miti_global}
\scriptsize
\setlength{\tabcolsep}{2.5pt}
\renewcommand{\arraystretch}{0.95}
\resizebox{\linewidth}{!}{
\begin{tabular}{@{}lccccc ccccc@{}}
\toprule
\multirow{2}{*}{\textbf{Method}} 
& \multicolumn{5}{c}{\textbf{Qwen3-8B}} 
& \multicolumn{5}{c}{\textbf{Llama-3.1-8B}} \\
\cmidrule(lr){2-6} \cmidrule(lr){7-11}
& \textbf{CultCT}$\uparrow$ 
& \textbf{SoftST}$\uparrow$ 
& \textbf{Partner}$\uparrow$ 
& \textbf{Empathy}$\uparrow$ 
& \textbf{Avg}$\uparrow$
& \textbf{CultCT}$\uparrow$ 
& \textbf{SoftST}$\uparrow$ 
& \textbf{Partner}$\uparrow$ 
& \textbf{Empathy}$\uparrow$ 
& \textbf{Avg}$\uparrow$ \\
\midrule
CoS 
& 1.31 & 1.41 & 1.41 & 1.76 & 1.47
& 1.24 & 1.28 & 1.24 & 1.69 & 1.36 \\
DIIR 
& 1.97 & 2.45 & 2.41 & 2.52 & 2.34
& 2.03 & 2.52 & 2.48 & 2.45 & 2.37 \\
CAMI-base 
& 1.59 & 2.00 & 2.00 & 2.03 & 1.91
& 1.55 & 2.00 & 1.93 & 1.90 & 1.84 \\
CAMI-full 
& 1.72 & 1.72 & 1.72 & 1.79 & 1.74
& 1.72 & 1.83 & 1.76 & 1.72 & 1.76 \\
CAMI-text 
& 1.83 & 1.97 & 1.86 & 1.83 & 1.87
& 2.07 & 2.21 & 2.14 & 2.14 & 2.14 \\
Real (human) 
& 3.38 & 4.10 & 4.00 & 3.93 & 3.85
& 3.38 & 4.10 & 4.00 & 3.93 & 3.85 \\
\midrule
\textbf{PUMA} 
& \textbf{3.59} & \textbf{4.59} & \textbf{4.66} & \textbf{4.62} & \textbf{4.37}
& \textbf{3.62} & \textbf{4.45} & \textbf{4.55} & \textbf{4.45} & \textbf{4.27} \\
\bottomrule
\end{tabular}
}
\vspace{-6pt}
\end{table}

\subsection{Ablation Study (RQ2)}
\label{sec:ablation}

Table~\ref{tab:ablation_dynamic} shows that all components contribute to dynamic dialogue control.
Starting from the full PUMA agent, removing belief, world-model context or EFE-based action selection each causes a comparable drop in Lift and Prep, while removing both the world model and action selection leads to the largest degradation.
Although \textit{"PUMA w/o WM input"} achieves a slightly higher TrigCov, this is because, without world-model-based future prediction, the planner is biased toward exploratory TrigCov rather than transition-aware user-state intervention.
This suggests that belief, the world model, and EFE-based action selection provide benefits for selecting state-appropriate actions and improving motivational outcomes.
\begin{table}[t]
\centering
\caption{Ablation study on CAMI. We remove belief, world-model input, and EFE-action selection to evaluate the contribution of each component; $\Delta$Lift is computed relative to the full PUMA model.}
\label{tab:ablation_dynamic}
\small
\setlength{\tabcolsep}{3pt}
\renewcommand{\arraystretch}{0.95}
\begin{tabular}{@{}lccccccc@{}}
\toprule
\textbf{Configuration} 
& \textbf{Prior} 
& \textbf{WM} 
& \textbf{Plan} 
& \textbf{Lift}$\uparrow$ 
& \textbf{Prep\%}$\uparrow$ 
& \textbf{TrigCov}$\uparrow$ 
& $\Delta$\textbf{Lift} \\
\midrule
CAMI-base 
& \xmark & \xmark & \xmark 
& 0.62 & 28.0\% & 22.0\% & --- \\

DIIR 
& partial & \xmark & \xmark 
& 1.07 & 52.0\% & 34.0\% & --- \\

\midrule
PUMA w/o WM+Plan 
& \cmark & \xmark & \xmark 
& 1.35 & 55.2\% & 54.0\% & $-$0.28 \\

PUMA w/o Prior Inf. 
& \xmark & \cmark & \cmark 
& 1.41 & 62.1\% & 56.6\% & $-$0.21 \\

PUMA w/o WM input 
& \cmark & \xmark$^\dagger$ & \cmark 
& 1.45 & 62.1\% & \textbf{63.8\%} & $-$0.17 \\

PUMA w/o EFE 
& \cmark & \cmark & \xmark 
& 1.41 & 62.1\% & 54.9\% & $-$0.21 \\

\midrule
\textbf{PUMA} 
& \cmark & \cmark & \cmark 
& \textbf{1.62} & \textbf{75.9\%} & 62.4\% & --- \\
\bottomrule
\end{tabular}

\vspace{2pt}
\footnotesize{$^\dagger$ The world model is not used as planner input, while EFE-based action selection remains enabled.}
\vspace{-8pt}
\end{table}
\subsection{In-depth Analysis (RQ3)}
\label{sec:analysis}

In this section, we further analyze why PUMA improves dynamic MI-coaching effectiveness.
Different from the main experiments, this section focuses on whether the proposed framework provides a reliable user-state mechanism for action selection.
We examine three aspects: current-state estimation, next-state prediction, and dynamic readiness progression.

\begin{figure}[H]
\centering
\begin{minipage}{0.95\linewidth}
\centering
\begin{subfigure}{0.46\linewidth}
    \centering
    \includegraphics[width=\linewidth]{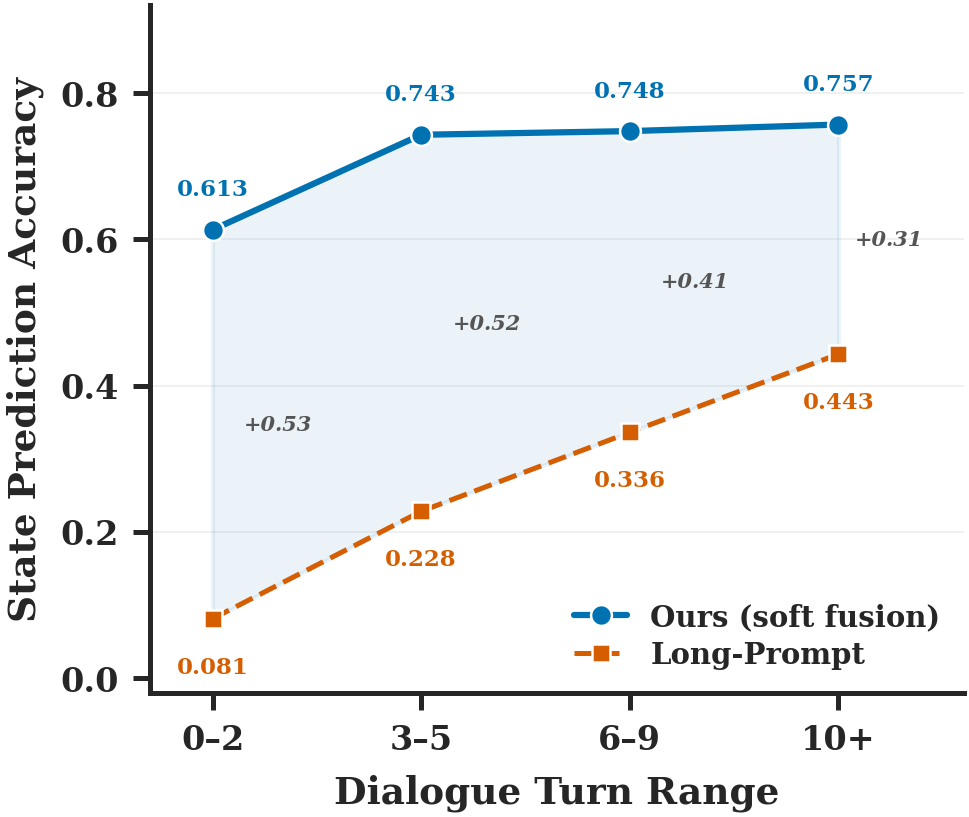}
    \caption{Next-state prediction accuracy.}
    \label{fig:next_state_tracking_curve}
\end{subfigure}
\hspace{0.06\linewidth}
\begin{subfigure}{0.46\linewidth}
    \centering
    \includegraphics[width=\linewidth]{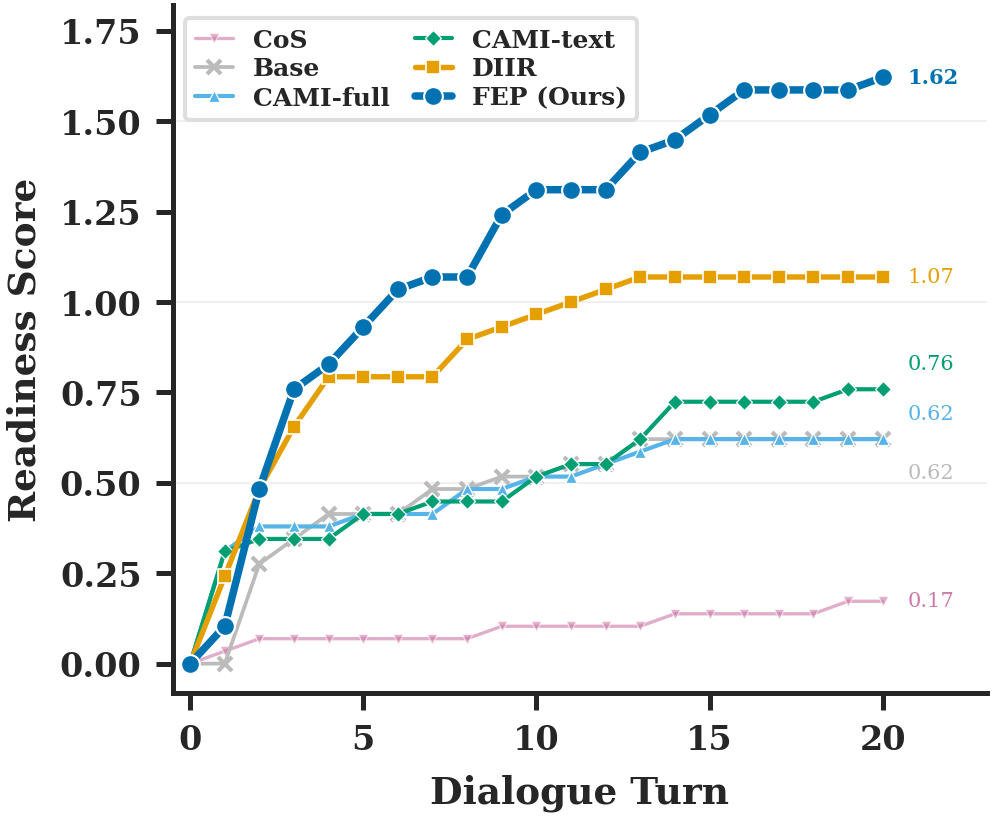}
    \caption{Dynamic readiness progression.}
    \label{fig:dynamic_readiness_curve}
\end{subfigure}
\end{minipage}
\caption{Temporal analysis of state prediction and readiness progression.}
\vspace{-1em}
\label{fig:state_tracking}
\end{figure}

\paragraph{State inference and temporal dynamics.}
Table~\ref{tab:static_main} shows that PUMA achieves the best performance on both Curr-Acc and Next-Acc, demonstrating that it can accurately estimate the user's current motivational state while predicting how the state may evolve next. 
These results suggest that the belief-world-model mechanism provides reliable state and transition information, offering a solid foundation for state-aware action selection.
Figure~\ref{fig:next_state_tracking_curve} further shows stable next-state tracking across dialogue stages, with accuracy increasing from 0.613 at turns 0--2 to 0.757 after turn 10, suggesting that accumulated interaction evidence leads to more reliable state estimates. 
Meanwhile, Figure~\ref{fig:dynamic_readiness_curve} shows that PUMA continues to improve client readiness throughout the interaction, whereas baselines tend to plateau after an initial increase, highlighting the benefit of state-aware action selection for long-horizon dialogue progression.

\paragraph{Case Study.}
We provide a qualitative case study illustrating how belief tracking supports dynamic action selection, with full interaction details reported in Appendix~\ref{app:case_study}.

\begin{table*}[t]
\centering
\small
\setlength{\tabcolsep}{3.5pt}
\begin{minipage}[t]{0.42\textwidth}
\vspace{0pt}
\centering
\caption{Current- and next-state accuracy.}
\label{tab:static_main}
\begin{tabular}{@{}lcc@{}}
\toprule
\textbf{Method} 
& \textbf{Curr-Acc}$\uparrow$ 
& \textbf{Next-Acc}$\uparrow$ \\
\midrule
CAMI-full 
& 0.504 & -- \\
Long-prompt 
& 0.667 & 0.231 \\
\midrule
\textbf{PUMA} 
& \textbf{0.689} 
& \textbf{0.717} \\
\bottomrule
\end{tabular}
\end{minipage}
\hfill
\begin{minipage}[t]{0.52\textwidth}
\vspace{0pt}
\centering
\caption{Generalization results on AnnoMI.}
\label{tab:annomi_generalization}
\begin{tabular}{@{}lccc@{}}
\toprule
\textbf{Method} 
& \textbf{Model} 
& \textbf{Curr-Acc}$\uparrow$ 
& \textbf{Next-Acc}$\uparrow$ \\
\midrule
Long-prompt 
& Qwen3-8B  
& 0.463 & 0.364 \\
Long-prompt 
& Qwen3-32B 
& 0.602 & 0.430 \\
\midrule
\textbf{PUMA} 
& Qwen3-8B 
& \textbf{0.639} 
& \textbf{0.532} \\
\bottomrule
\end{tabular}
\end{minipage}

\vspace{-0.8em}
\end{table*}

\subsection{Cross-Dataset Generalizability}
\label{sec:generalizability_summary}

We further test whether the proposed user-state modeling framework generalizes beyond CAMI through a cross-dataset study on AnnoMI~\citep{wu2022anno}.
The detailed setup is provided in Appendix~\ref{sec:generalizability}.
As shown in Table~\ref{tab:annomi_generalization}, PUMA with Qwen3-8B achieves higher Curr-Acc and Next-Acc than the long-prompt baseline with Qwen3-32B, despite using a smaller backbone.
This result highlights the strength and scalability of the proposed user-state modeling framework: even with a smaller backbone, PUMA achieves stronger current-state estimation and next-state prediction than a much larger long-prompt model.
These results suggest that the framework can transfer across different user-state annotation schemes.


\section{Conclusion}
\label{sec:conclusion}

We presented \textbf{PUMA}, an FEP-grounded framework for personalized dialogue. 
Instead of treating personalization as passive reuse of explicit user histories, our framework models the user as a partially observed dynamical system. This goes beyond static state estimation by explicitly modeling state transitions induced by dialogue actions.
By maintaining a belief over the latent user state, refining a world model, and selecting actions through expected free energy, PUMA enables the system to reason about both the current interpretation of the user and the future consequences of its responses. 

Experiments in healthcare-oriented motivational interviewing settings show that PUMA improves dynamic counseling effectiveness, maintains strong counselor-side response quality, and provides reliable estimation.
Ablation and cross-dataset results further show that belief updating, world-model refinement, and expected-free-energy-based action selection each contribute to robust long-horizon personalization.
These findings suggest that personalized dialogue systems can move beyond memory retrieval toward active, state-aware interaction control.


\bibliographystyle{plainnat}
\bibliography{references}

\newpage
\appendix

\section{Evaluation Metrics}
\label{app:metrics}

We evaluate the framework from three perspectives: static state inference, dynamic counseling effectiveness, and counselor-side response quality.

\paragraph{Dynamic counseling effectiveness.}
\textbf{Stage Lift} measures the average ordinal improvement from the initial to the final motivational stage.
\textbf{Prep Rate} is the percentage of profiles that reach the preparation stage.
\textbf{Trigger Coverage} measures the fraction of profile-specific motivational triggers addressed by the counselor via semantic matching.
\textbf{Avg Turns} reports the average number of turns used to complete the session, where lower values indicate higher efficiency when stage progression is strong.

\paragraph{Counselor-side response quality.}
We use four \textbf{MITI Global Scores}~\citep{yang2025cami}.
\textbf{Cultivating Change Talk} measures the extent to which the counselor encourages client language favoring change.
\textbf{Softening Sustain Talk} measures the extent to which the counselor avoids amplifying resistance.
\textbf{Partnership} measures whether the counselor conveys that expertise and choice reside within the client.
\textbf{Empathy} measures whether the counselor understands and reflects the client's perspective.
Each score is rated on a 1--5 Likert scale.
We report the average of the four scores as \textbf{MITI-Avg}.

\paragraph{Static state inference.}
\textbf{Curr-Acc} measures the proportion of turns where the predicted current motivational state matches the gold label.
\textbf{Next-Acc} measures the accuracy of predicting the client's motivational state at the following turn.

\section{Technical Appendices}
\label{sec:appendix}
\subsection{Derivation of the Posterior Inference Objective}
\label{app:posterior_derivation}
We derive the posterior inference objective from the standard free-energy construction.
At turn \(t\), the user world model factorizes the joint distribution of the current observation and latent user state as
\begin{equation}
p_\theta(o_t, s_t \mid s_{t-1}, a_{t-1})
=
p_\theta(o_t \mid s_t)
p_\theta(s_t \mid s_{t-1}, a_{t-1}).
\label{eq:app_world_model}
\end{equation}
For notational simplicity, we omit the conditioning on previous states and actions in the following derivation, and write the current-turn joint distribution as \(p_\theta(o_t,s_t)\).

Following the Free Energy Principle, the key goal can be formulated as reducing the surprise of observations under the agent's generative model.
For the current user observation \(o_t\), this surprise is given by
\begin{equation}
-\log p_\theta(o_t).
\label{eq:app_surprise}
\end{equation}
However, evaluating this quantity requires computing the marginal likelihood, or model evidence, of the observation:
\begin{equation}
p_\theta(o_t)
=
\int p_\theta(o_t,s_t)\,ds_t.
\label{eq:app_evidence}
\end{equation}
This term marginalizes over all possible latent user states that could have generated the current observation.
Since the latent state space is generally large or continuous, directly computing \(p_\theta(o_t)\), and hence the surprise \(-\log p_\theta(o_t)\), is generally intractable.

To obtain a tractable objective, we introduce an auxiliary distribution \(q(s_t)\) over the latent state.
This introduction does not change the marginal likelihood, since
\begin{align}
p_\theta(o_t)
&=
\int p_\theta(o_t,s_t)\,ds_t \\
&=
\int
q(s_t)
\frac{
p_\theta(o_t,s_t)
}{
q(s_t)
}
\,ds_t \\
&=
\mathbb{E}_{q(s_t)}
\left[
\frac{
p_\theta(o_t,s_t)
}{
q(s_t)
}
\right].
\label{eq:app_q_identity}
\end{align}

Substituting Eq.~\ref{eq:app_q_identity} into the surprise gives
\begin{align}
-\log p_\theta(o_t)
&=
-\log
\mathbb{E}_{q(s_t)}
\left[
\frac{
p_\theta(o_t,s_t)
}{
q(s_t)
}
\right].
\label{eq:app_surprise_expectation}
\end{align}
Since \(-\log(\cdot)\) is convex, Jensen's inequality gives
\begin{align}
-\log p_\theta(o_t)
&\leq
\mathbb{E}_{q(s_t)}
\left[
-\log
\frac{
p_\theta(o_t,s_t)
}{
q(s_t)
}
\right] \\
&=
\mathbb{E}_{q(s_t)}
\left[
\log q(s_t)
-
\log p_\theta(o_t,s_t)
\right].
\label{eq:app_jensen_bound}
\end{align}
The right-hand side is the variational free energy:
\begin{equation}
\mathcal{F}_t(q)
=
\mathbb{E}_{q(s_t)}
\left[
\log q(s_t)
-
\log p_\theta(o_t,s_t)
\right].
\label{eq:app_vfe_definition}
\end{equation}
Thus, variational free energy provides an upper bound on the surprise:
\begin{equation}
-\log p_\theta(o_t)
\leq
\mathcal{F}_t(q).
\label{eq:app_free_energy_bound}
\end{equation}
Following the Free Energy Principle, the agent avoids surprising observations by minimizing this upper bound.
Since the true surprise \(-\log p_\theta(o_t)\) is difficult to compute directly, the agent instead optimizes the tractable objective \(\mathcal{F}_t(q)\) with respect to its belief \(q(s_t)\):
\begin{equation}
q^\star(s_t)
=
\arg\min_q \mathcal{F}_t(q).
\label{eq:app_q_optimization}
\end{equation}

Using the current-turn factorization
\begin{equation}
p_\theta(o_t,s_t)
=
p_\theta(o_t\mid s_t)p_\theta(s_t),
\label{eq:app_current_joint}
\end{equation}
we can rewrite the variational free energy as
\begin{align}
\mathcal{F}_t(q)
&=
\mathbb{E}_{q(s_t)}
\left[
\log q(s_t)
-
\log p_\theta(s_t)
-
\log p_\theta(o_t\mid s_t)
\right] \\
&=
D_{\mathrm{KL}}
\left(
q(s_t)\,\|\,p_\theta(s_t)
\right)
-
\mathbb{E}_{q(s_t)}
\left[
\log p_\theta(o_t\mid s_t)
\right].
\label{eq:app_dialogue_vfe}
\end{align}
This decomposition gives the posterior update objective used in our framework.
The first term keeps the updated belief \(q(s_t)\) close to the prior \(p_\theta(s_t)\), which encodes the expected user state before observing \(o_t\).
The second term encourages \(q(s_t)\) to place probability mass on latent states that explain the current observation through the likelihood \(p_\theta(o_t\mid s_t)\).
Therefore, minimizing \(\mathcal{F}_t(q)\) updates the system's belief by balancing prior consistency and observation fit.

To show that this optimization corresponds to posterior inference, we can also rewrite the joint distribution as
\begin{equation}
p_\theta(o_t,s_t)
=
p_\theta(s_t\mid o_t)p_\theta(o_t).
\label{eq:app_posterior_factorization}
\end{equation}
Substituting this into Eq.~\ref{eq:app_vfe_definition} yields
\begin{align}
\mathcal{F}_t(q)
&=
\mathbb{E}_{q(s_t)}
\left[
\log q(s_t)
-
\log p_\theta(s_t\mid o_t)
-
\log p_\theta(o_t)
\right] \\
&=
D_{\mathrm{KL}}
\left(
q(s_t)\,\|\,p_\theta(s_t\mid o_t)
\right)
-
\log p_\theta(o_t).
\label{eq:app_vfe_posterior_equivalence}
\end{align}
The second term \(-\log p_\theta(o_t)\) does not depend on \(q(s_t)\).
Therefore, minimizing \(\mathcal{F}_t(q)\) with respect to \(q(s_t)\) is equivalent to minimizing
\begin{equation}
D_{\mathrm{KL}}
\left(
q(s_t)\,\|\,p_\theta(s_t\mid o_t)
\right).
\end{equation}

Thus, \(q(s_t)\) becomes an approximate posterior after optimization, rather than being assumed to equal the posterior from the beginning.

\subsection{Derivation of the World Model Learning Objective}
\label{app:wm_derivation}

We derive the world model learning objective from the per-turn posterior
inference objective established in Appendix~\ref{app:posterior_derivation}.
Starting from the variational free energy at turn \(t\),
\begin{equation}
\mathcal{F}_t(q,\theta)
=
D_{\mathrm{KL}}
\left(
q(s_t)\,\|\,\tilde p_\theta(s_t)
\right)
-
\mathbb{E}_{q(s_t)}
\left[
\log p_\theta(o_t \mid s_t)
\right],
\label{eq:app_wm_vfe}
\end{equation}
we note that this objective depends on two sets of parameters: the
variational parameters \(\phi\) of the belief \(q_\phi(s_t)\), and the
generative parameters \(\theta\) of the world model.
Posterior inference (Eq.~\ref{eq:posterior_inference}) optimizes
\(\mathcal{F}_t\) with respect to \(\phi\) at a single turn.
World model learning instead optimizes the same free energy with respect to
\(\theta\), accumulated over the dialogue.

Aggregating Eq.~\ref{eq:app_wm_vfe} along the trajectory yields
\begin{equation}
\mathcal{F}(\theta)
=
\sum_{t=1}^{T}
\mathcal{F}_t(q_\phi,\theta).
\label{eq:app_traj_sum}
\end{equation}
We now identify the \(\theta\)-dependent components in
Eq.~\ref{eq:app_traj_sum}.

\paragraph{Observation term.}
The likelihood term in Eq.~\ref{eq:app_wm_vfe} contributes, at each turn,
\begin{equation}
-\mathbb{E}_{q_\phi(s_t)}
\left[
\log p_\theta(o_t \mid s_t)
\right],
\label{eq:app_obs_term}
\end{equation}
which trains the observation model to assign high likelihood to the actual
user utterance \(o_t\) under the inferred latent state.

\paragraph{Transition term.}
The KL term in Eq.~\ref{eq:app_wm_vfe} depends on \(\theta\) only through
the predictive prior
\begin{equation}
\tilde p_\theta(s_t)
=
\!\int\! p_\theta(s_t\mid s_{t-1},a_{t-1})\,
q_\phi(s_{t-1})\,ds_{t-1},
\label{eq:app_predictive_prior}
\end{equation}
which is determined by the action-conditioned transition model
\(p_\theta(s_t\mid s_{t-1},a_{t-1})\).
Reindexing this contribution from \(t\) to \(t+1\) so that it expresses how
state \(s_{t+1}\) is predicted from \(s_t\) and \(a_t\), the
\(\theta\)-dependent part of the prior takes the form
\begin{equation}
-\mathbb{E}_{q_\phi(s_t)}
\left[
\log p_\theta(s_{t+1}\mid s_t, a_t)
\right],
\quad t=1,\ldots,T-1,
\label{eq:app_trans_term}
\end{equation}
which trains the transition model to predict the next inferred user state
from the current state and system action.

\paragraph{Combined objective.}
Summing Eqs.~\ref{eq:app_obs_term} and~\ref{eq:app_trans_term} over the
trajectory and introducing a coefficient \(\lambda\) to balance the relative
scales of the observation and transition losses gives
\begin{equation}
\mathcal{L}_{\mathrm{WM}}
=
\sum_{t=1}^{T}
\mathbb{E}_{q_\phi(s_t)}
\left[
-\log p_\theta(o_t \mid s_t)
\right]
+
\lambda
\sum_{t=1}^{T-1}
\mathbb{E}_{q_\phi(s_t)}
\left[
-\log p_\theta(s_{t+1} \mid s_t, a_t)
\right],
\label{eq:app_wm_objective}
\end{equation}
which is exactly Eq.~\ref{eq:wm_learning}.
Thus, posterior inference and world model learning optimize the same per-turn
variational free energy with respect to \(\phi\) and \(\theta\) respectively,
with the latter aggregated along the dialogue trajectory.

\subsection{Derivation of the Expected Free Energy Objective}
\label{app:efe_derivation}

The posterior inference objective in Appendix~\ref{app:posterior_derivation} evaluates the variational free energy at the current turn, where the observation \(o_t\) is already available.
For action selection, however, the agent must reason about a future turn \(t+1\), at which the next observation \(o_{t+1}\) has not yet been received.
We extend the free energy into an \emph{expected free energy} that scores each candidate action \(a\), starting from the equivalent form of the variational free energy obtained in Eq.~\ref{eq:app_vfe_posterior_equivalence}:
\begin{equation}
\mathcal{F}_t(q)
=
D_{\mathrm{KL}}
\left(
q(s_t)\,\|\,p_\theta(s_t\mid o_t)
\right)
-
\log p_\theta(o_t).
\label{eq:app_efe_starting_point}
\end{equation}

Since \(-\log p_\theta(o_t)\) does not depend on \(s_t\), it can be absorbed into the expectation under \(q(s_t)\), yielding the joint-surprise form
\begin{align}
\mathcal{F}_t(q)
&=
\mathbb{E}_{q(s_t)}
\left[
\log q(s_t)
-
\log p_\theta(s_t\mid o_t)
-
\log p_\theta(o_t)
\right] \\
&=
\mathbb{E}_{q(s_t)}
\left[
\log q(s_t)
-
\log p_\theta(s_t,o_t)
\right].
\label{eq:app_vfe_joint_form}
\end{align}

To plan an action \(a\) at turn \(t\), we evaluate the analogous quantity at turn \(t+1\) under three modifications.
\textit{(i)}~The future observation \(o_{t+1}\) is unknown, so we take an additional expectation over the predicted observation distribution implied by the world model,
\begin{equation}
q(o_{t+1}\mid a)
=
\int
q(s_{t+1}\mid a)\,
p_\theta(o_{t+1}\mid s_{t+1})\,
ds_{t+1}.
\label{eq:app_predicted_obs}
\end{equation}
\textit{(ii)}~The current belief \(q(s_t)\) is replaced by the prior-predictive belief \(q(s_{t+1}\mid a)\), obtained by propagating \(q(s_t)\) one step forward through the user transition model conditional on \(a\).
\textit{(iii)}~The model evidence \(p_\theta(o_{t+1})\), which is a neutral marginal in the inference setting, is replaced by a preference distribution \(p_{\mathrm{pref}}(o_{t+1})\) encoding the desired future outcomes.
Equivalently, the joint distribution \(p_\theta(s_{t+1},o_{t+1}\mid a)\) is replaced by the preference-biased joint
\begin{equation}
\tilde{p}(s_{t+1},o_{t+1}\mid a)
=
p_\theta(s_{t+1}\mid o_{t+1},a)\,
p_{\mathrm{pref}}(o_{t+1}).
\label{eq:app_efe_biased_joint}
\end{equation}
This substitution is the central modeling step of active inference: it turns the inference objective into a planning objective by aligning low surprise with desired outcomes rather than with predictive accuracy alone.

Applying these three modifications to Eq.~\ref{eq:app_vfe_joint_form}, we define the expected free energy under action \(a\) as
\begin{align}
G(a)
&=
\mathbb{E}_{q(s_{t+1},o_{t+1}\mid a)}
\left[
\log q(s_{t+1}\mid a)
-
\log \tilde{p}(s_{t+1},o_{t+1}\mid a)
\right] \\
&=
\mathbb{E}_{q(s_{t+1},o_{t+1}\mid a)}
\left[
\log q(s_{t+1}\mid a)
-
\log p_\theta(s_{t+1}\mid o_{t+1},a)
-
\log p_{\mathrm{pref}}(o_{t+1})
\right].
\label{eq:app_efe_definition}
\end{align}

The term \(p_\theta(s_{t+1}\mid o_{t+1},a)\) is the true posterior over the future user state, which is intractable for the same reason \(p_\theta(s_t\mid o_t)\) is intractable in the current-turn case.
Following the same variational principle, we approximate it by the variational posterior \(q(s_{t+1}\mid o_{t+1},a)\):
\begin{align}
G(a)
&\approx
\mathbb{E}_{q(s_{t+1},o_{t+1}\mid a)}
\left[
\log q(s_{t+1}\mid a)
-
\log q(s_{t+1}\mid o_{t+1},a)
\right] \notag\\
&\quad
+
\mathbb{E}_{q(o_{t+1}\mid a)}
\left[
-\log p_{\mathrm{pref}}(o_{t+1})
\right].
\label{eq:app_efe_after_approx}
\end{align}

The first term in Eq.~\ref{eq:app_efe_after_approx} admits an information-theoretic interpretation.
Factoring the joint expectation as \(q(s_{t+1},o_{t+1}\mid a)=q(o_{t+1}\mid a)\,q(s_{t+1}\mid o_{t+1},a)\),
\begin{align}
&\mathbb{E}_{q(s_{t+1},o_{t+1}\mid a)}
\left[
\log q(s_{t+1}\mid a)
-
\log q(s_{t+1}\mid o_{t+1},a)
\right] \notag\\
&=
-\mathbb{E}_{q(o_{t+1}\mid a)}
\left[
D_{\mathrm{KL}}
\left(
q(s_{t+1}\mid o_{t+1},a)\,\|\,q(s_{t+1}\mid a)
\right)
\right] \\
&=
-I(s_{t+1};o_{t+1}\mid a),
\label{eq:app_epistemic_mi}
\end{align}
where \(I(s_{t+1};o_{t+1}\mid a)\) is the mutual information between the future user state and the future observation conditional on action \(a\).
This quantity, the \emph{epistemic value} of the action, measures how much the agent expects to learn about \(s_{t+1}\) by observing the user response \(o_{t+1}\) after taking action \(a\).
Equivalently, the mutual information decomposes into a difference of entropies:
\begin{equation}
I(s_{t+1};o_{t+1}\mid a)
=
H\!\left(q(s_{t+1}\mid a)\right)
-
\mathbb{E}_{q(o_{t+1}\mid a)}
\left[
H\!\left(q(s_{t+1}\mid o_{t+1},a)\right)
\right].
\label{eq:app_mi_entropy_form}
\end{equation}

Substituting Eq.~\ref{eq:app_mi_entropy_form} back into Eq.~\ref{eq:app_efe_after_approx} gives
\begin{align}
G(a)
&\approx
\mathbb{E}_{q(o_{t+1}\mid a)}
\left[
H\!\left(q(s_{t+1}\mid o_{t+1},a)\right)
\right]
-
H\!\left(q(s_{t+1}\mid a)\right) \notag\\
&\quad
+
\mathbb{E}_{q(s_{t+1},o_{t+1}\mid a)}
\left[
-\log p_{\mathrm{pref}}(o_{t+1})
\right].
\label{eq:app_efe_three_terms}
\end{align}

Action selection only requires \(\arg\min_{a\in\mathcal{A}} G(a)\), so any term that is approximately invariant across candidate actions does not affect the decision.
The prior-predictive entropy \(H\!\left(q(s_{t+1}\mid a)\right)\) is dominated by the current belief \(q(s_t)\) propagated through the transition model and varies negligibly across candidate actions compared with the data-driven posterior entropy and the preference-weighted log-likelihood.
Treating it as a constant absorbs it into the \(\arg\min\) and yields the action-selection objective used in Section~\ref{sec:action_selection}:
\begin{equation}
G(a)
=
\underbrace{
\mathbb{E}_{q(o_{t+1}\mid a)}
\left[
H\!\left(q(s_{t+1}\mid o_{t+1},a)\right)
\right]
}_{\text{exploration term}}
+
\underbrace{
\mathbb{E}_{q(s_{t+1},o_{t+1}\mid a)}
\left[
-\log p_{\mathrm{pref}}(o_{t+1})
\right]
}_{\text{exploitation term}},
\label{eq:app_efe_final}
\end{equation}
with the selected action \(a_t=\arg\min_{a\in\mathcal{A}} G(a)\).

The exploration term is the expected entropy of the agent's belief about the next user state \emph{after} taking action \(a\) and observing the resulting response, so minimizing it favors actions whose predicted observations are most informative about the user.
The exploitation term is the expected negative log-preference of the predicted observation, so minimizing it favors actions whose consequences best match the desired future outcomes.
Together, these two terms realize the active-inference principle that an agent should both reduce uncertainty about the user and steer interactions toward preferred states.
\section{Hyperparameter Settings}
\label{app:hyperparams}

This appendix summarizes the hyperparameter settings used in our experiments.
Unless otherwise specified, all experiments use the default values reported in Table~\ref{tab:general_hyperparams}.
The core PUMA hyperparameters are shared by both the static and dynamic evaluation tracks, although they are passed through different configuration interfaces: the static track uses \texttt{config.py}, while the dynamic track uses script-level \texttt{argparse} arguments or internal default dictionaries.
In large-scale runs, Slurm scripts may override these default values.

\begin{table}[t]
\centering
\small
\setlength{\tabcolsep}{4pt}
\renewcommand{\arraystretch}{1.15}
\caption{General hyperparameters used in PUMA.}
\label{tab:general_hyperparams}
\begin{tabularx}{\linewidth}{L{0.31\linewidth} L{0.18\linewidth} Y}
\toprule
\textbf{Hyperparameter} & \textbf{Default Value} & \textbf{Description} \\
\midrule
\makecell[l]{\texttt{relevance\_memory}\\\texttt{\_number}}
& 1
& Number of retrieved relevant memories. \\

\makecell[l]{\texttt{context\_memory}\\\texttt{\_number}}
& 30
& Maximum number of contextual memories. \\

\texttt{dist\_thres}
& 1.5
& Distance threshold for relevance-based retrieval. \\

\makecell[l]{\texttt{persona\_history}\\\texttt{\_max}}
& 6
& Maximum history turns for long-term persona modeling. \\

\makecell[l]{\texttt{persona\_short\_term}\\\texttt{\_window\_turns}}
& 4
& Short-term window size for \(q\)-state inference. \\

\makecell[l]{\texttt{persona\_long\_term}\\\texttt{\_summary\_turns}}
& 12
& Number of turns used for long-term \(p\)-state summary update. \\

\makecell[l]{\texttt{fep\_lambda\_e} /\\\texttt{fep\_lambda\_p}}
& 0.4 / 0.6
& Epistemic and pragmatic weights in EFE-based action selection. \\

\makecell[l]{\texttt{fep\_soft\_fusion}\\\texttt{\_beta}}
& 0.35
& Weight of the planner prior in soft fusion. \\

\makecell[l]{\texttt{max\_output}\\\texttt{\_length}}
& 1024
& Maximum number of generated LLM tokens. \\

\makecell[l]{\texttt{cami\_warmup\_ratio} /\\\texttt{cami\_min\_eval\_turns}}
& 0.5 / 3
& Warm-up and evaluation split for CAMI data. \\

\texttt{max\_turns}
& 12
& Maximum number of counselor turns in \texttt{interactive\_cami\_text\_eval}. \\
\bottomrule
\end{tabularx}
\end{table}

\begin{table}[t]
\centering
\small
\setlength{\tabcolsep}{4pt}
\renewcommand{\arraystretch}{1.15}
\caption{Hyperparameters for \texttt{eval\_sim.py}.}
\label{tab:simple_sim_hyperparams}
\begin{tabularx}{\linewidth}{L{0.31\linewidth} L{0.24\linewidth} Y}
\toprule
\textbf{Hyperparameter} & \textbf{Default Value} & \textbf{Description} \\
\midrule
\texttt{max\_turns}
& 20
& Maximum number of dialogue turns. \\

\texttt{seed}
& 42
& Random seed. \\

\texttt{test\_num}
& 5
& Number of evaluated profiles by default. \\

\makecell[l]{\texttt{fep\_lambda\_e} /\\\texttt{fep\_lambda\_p}}
& 0.4 / 0.6
& Epistemic and pragmatic weights in EFE-based action selection. \\

\texttt{disable\_planner}
& \texttt{False}
& Whether to disable the transition planner. \\

\makecell[l]{\texttt{observation\_focused}\\\texttt{\_state}}
& \texttt{True}
& Whether to use the observation-focused state head. \\

\texttt{fep\_efe\_action}
& \texttt{True}
& Whether to use EFE-based action selection. \\

\makecell[l]{\texttt{cami\_preparation}\\\texttt{\_calibration}}
& \texttt{True}
& Whether to apply preparation-stage calibration. \\

\texttt{doctor\_modes}
& \makecell[l]{\texttt{fep}, \texttt{cami\_text},\\
\texttt{base\_counselor}, \texttt{diir},\\
\texttt{cos}, \texttt{cami\_full}}
& Compared counselor agents. \\
\bottomrule
\end{tabularx}
\end{table}

\begin{table}[t]
\centering
\small
\setlength{\tabcolsep}{4pt}
\renewcommand{\arraystretch}{1.15}
\caption{Hyperparameters for \texttt{eval\_patient\_simulator.py}.}
\label{tab:patient_sim_hyperparams}
\begin{tabularx}{\linewidth}{L{0.31\linewidth} L{0.18\linewidth} Y}
\toprule
\textbf{Hyperparameter} & \textbf{Default Value} & \textbf{Description} \\
\midrule
\texttt{max\_turns}
& 20
& Maximum number of dialogue turns. \\

\texttt{seed}
& 42
& Random seed. \\

\texttt{judge\_model}
& \texttt{Qwen3-32B}
& LLM judge used for simulator evaluation. \\

\texttt{load\_in\_4bit}
& \texttt{False}
& Whether to enable 4-bit quantized loading. \\
\bottomrule
\end{tabularx}
\end{table}
\paragraph{Shared PUMA Parameters.}
The two evaluation tracks share the same core PUMA parameters.
Specifically, \texttt{fep\_lambda\_e}=0.4 and \texttt{fep\_lambda\_p}=0.6 control the trade-off between epistemic value and pragmatic value in expected-free-energy action selection.
The soft-fusion coefficient \texttt{fep\_soft\_fusion\_beta}=0.35 controls the contribution of the planner prior when combining model-based predictions with observation-driven state inference.
These parameters are kept fixed across the main experiments unless explicitly changed in ablation or sensitivity analysis.
\section{Client Simulator Details and Validation}
\label{app:simulator_details}

This appendix describes the design and validation of \textbf{DynPatient} , the client simulator used for simulator-based dynamic evaluation.
The simulator provides a controlled environment for comparing counselor policies under the same client profiles and initial conditions.
It is used only for dynamic evaluation, while static state inference and next-state prediction are evaluated directly against gold annotated turns.

\subsection{Simulator Design}
\label{app:simulator_design}

\paragraph{Design goals.}
DynPatient is a profile-grounded and LLM-driven client simulator.
Given a counselor utterance, the simulator first determines whether the response addresses profile-specific motivational triggers, then updates an internal readiness variable, performs stage transition, selects a client action, and finally generates a natural-language client response.
The design follows three principles.
First, the simulator should implement a learnable state machine, where client-state transitions are driven by counselor behavior rather than random sampling.
Second, it should be profile-faithful, with each simulated client grounded in the corresponding CAMI profile, including beliefs, motivations, acceptable plans, personas, and behavior descriptions.
Third, it should be reproducible, using deterministic greedy decoding and fixed random seeds.

\paragraph{State space.}
The simulator follows the Transtheoretical Model stage space used in CAMI:
\begin{equation}
\mathcal{S}
=
\{
\textit{precontemplation},
\textit{contemplation},
\textit{preparation}
\}.
\end{equation}
Each episode starts from the initial stage given by the corresponding CAMI profile.
The dynamic evaluation measures whether a counselor can move the simulated client toward \emph{preparation} within a fixed turn budget.

\paragraph{Profile-grounded triggers.}
For each client profile, we construct a set of motivational triggers from the original profile sentences.
We use sentences from \emph{beliefs}, \emph{motivations}, and \emph{acceptable plans} as profile-specific anchors.
Each trigger is represented as a tuple \((c_i, z_i, b_i)\), where \(c_i\) denotes the source category, \(z_i\) is the trigger sentence, and \(b_i\) is its discovery bonus.
Personas are used as background context for response generation, but are not included in the trigger set.

\begin{table}[t]
\centering
\caption{Trigger construction in DynPatient.}
\label{tab:trigger_construction}
\begin{tabular}{lcc}
\toprule
\textbf{Source} & \textbf{Length filter} & \textbf{Discovery bonus} \\
\midrule
Beliefs & len \(> 10\) & 0.2 \\
Motivation & len \(> 20\) & 0.4 \\
Acceptable Plans & len \(> 10\) & 0.5 \\
\bottomrule
\end{tabular}
\end{table}

\paragraph{Semantic trigger matching.}
Given the counselor utterance \(u_t\), the simulator computes semantic similarity between \(u_t\) and each profile trigger.
We encode both trigger sentences and counselor utterances with a shared SentenceTransformer encoder and use cosine similarity for matching:
\begin{equation}
\mathcal{M}_t
=
\left\{
i :
\cos(\mathbf{e}_{z_i}, \mathbf{e}_{u_t}) \geq \tau
\right\},
\qquad
\tau = 0.45.
\end{equation}
Here, \(\mathcal{M}_t\) denotes the set of triggers matched at turn \(t\).
The threshold is set to avoid overly loose matches, so counselor responses must semantically address profile-specific content to affect the client's readiness.

\paragraph{Readiness dynamics.}
The simulator maintains a continuous readiness variable \(r_t \in \mathbb{R}\), initialized as \(r_0 = 0\).
At each turn, readiness is updated as:
\begin{equation}
r_{t+1}
=
r_t
+
\underbrace{
\mathbb{E}[\Delta r \mid s_t, a_t] \cdot g_t
}_{\text{action-driven contribution}}
+
\underbrace{
\sum_{i \in \mathcal{N}_t} b_i
}_{\text{discovery bonus}},
\label{eq:sim_readiness_update}
\end{equation}
where \(s_t\) is the current client stage, \(a_t\) is the selected client action, \(g_t\) is a content gate, and \(\mathcal{N}_t\) is the set of newly discovered triggers at turn \(t\).

The expected readiness increment is estimated from CAMI annotations:
\begin{equation}
\mathbb{E}[\Delta r \mid s, a]
=
\sum_{tt \in \{c,n,su\}}
P(tt \mid s, a) w_{tt},
\end{equation}
where \(tt\) denotes client talk type, including change talk \(c\), neutral talk \(n\), and sustain talk \(su\).
We use \(w_c = 1.0\), \(w_n = 0.3\), and \(w_{su} = -1.0\).
Only counselor-client turn pairs with sufficient support are used to estimate \(P(tt \mid s,a)\).

\paragraph{Content gate.}
The content gate prevents generic counselor responses from increasing readiness without addressing profile-specific content.
It is defined as:
\begin{equation}
g_t =
\begin{cases}
0.1 + 0.9 \cdot \rho_t \cdot \delta_t,
& \text{if } |\mathcal{M}_t| > 0, \\
0.1,
& \text{otherwise},
\end{cases}
\end{equation}
where \(\rho_t = \max_{i \in \mathcal{M}_t} \cos(\mathbf{e}_{z_i}, \mathbf{e}_{u_t})\) is the highest trigger-matching score.
The decay term \(\delta_t = 0.5^{h_t - 1}\) penalizes repeated matching of the same trigger, where \(h_t\) is the minimum previous hit count among the matched triggers.
Thus, repeated use of the same profile cue yields diminishing gains and encourages the counselor to discover more client-specific concerns.

\paragraph{Stage transition.}
The simulator uses two transition rules corresponding to the ordinal structure of the TTM stages \citep{prochaska2008initial}, including precontemplation, contemplation and preparation. 
The transition from \emph{precontemplation} to \emph{contemplation} is driven by trigger coverage:
\begin{equation}
s_{t+1} = \textit{contemplation}
\quad \text{iff} \quad
\frac{|\mathcal{D}_t|}{|\mathcal{T}|}
\geq
\theta_{\mathrm{cov}},
\qquad
\theta_{\mathrm{cov}} = 0.3,
\end{equation}
where \(\mathcal{D}_t\) is the set of discovered triggers and \(\mathcal{T}\) is the full trigger set for the profile.
When this transition occurs, readiness is reset to zero, because entering contemplation means that the client has started to recognize the problem rather than accumulated commitment to a concrete plan.

The transition from \emph{contemplation} to \emph{preparation} is driven by readiness:
\begin{equation}
s_{t+1} = \textit{preparation}
\quad \text{iff} \quad
r_t \geq \theta_{\mathrm{C}\rightarrow \mathrm{P}}^{(p)}.
\end{equation}
The threshold \(\theta_{\mathrm{C}\rightarrow \mathrm{P}}^{(p)}\) is calibrated separately for each profile by replaying the original CAMI trajectory and recording the readiness level at the observed transition point.
Profiles without a calibrated threshold use a default value of \(0.5\).

\paragraph{Client action selection.}
After stage transition, the simulator selects a client action.
To combine profile-specific behavior with population-level priors, we estimate a Dirichlet-smoothed action distribution:
\begin{equation}
P_p(a \mid s)
=
\frac{
n_{p,s,a} + \alpha P_{\mathrm{pop}}(a \mid s)
}{
n_{p,s,\cdot} + \alpha
},
\qquad
\alpha = 5.0,
\end{equation}
where \(n_{p,s,a}\) is the number of times client profile \(p\) takes action \(a\) at stage \(s\), and \(P_{\mathrm{pop}}(a \mid s)\) is the population-level action prior.
This distribution is provided to the LLM as a soft behavioral prior rather than used as a hard sampler.
The LLM receives the current stage, the profile-level action distribution, recent dialogue context, the counselor utterance, matched triggers, and persona information, and then selects one action from the stage-specific action set.
If action parsing fails, the simulator falls back to the highest-probability action under \(P_p(a \mid s)\).

\paragraph{Client response generation.}
Finally, the simulator generates the next client utterance using an LLM conditioned on the profile, recent dialogue history, current counselor utterance, selected client action, matched triggers, and current stage.
The prompt constrains the response to be short, first-person, conversational, and consistent with the client profile.
The generated response is not allowed to explicitly mention internal stage labels, therapy techniques, or simulator instructions.
This design encourages naturalistic client behavior while keeping the response grounded in the profile and the updated motivational state.
The complete prompt templates are provided in Appendix~\ref{app:prompts_sim}.

\subsection{Simulator Validation}
\label{app:simulator_validation}

\paragraph{Validation setup.}
Since the dynamic evaluation depends on simulated client responses, we validate DynPatient against representative client simulator baselines on the same CAMI profiles.
We compare against profile-based, profile-plus-action-based, CAMI dynamic , and Yang et al. simulators.\citep{yang2025consistent}
The validation evaluates both response-level quality and behavioral distribution matching.

\paragraph{Metrics.}
We use four simulator fidelity metrics.
\textbf{Consistency} measures whether the simulated client remains faithful to the assigned profile and motivational state.
\textbf{Realism} measures whether the generated utterances are natural and plausible as client responses in MI counseling.
Both scores are rated on a 1--5 scale.
\textbf{Quality} is the average of Consistency and Realism.
\textbf{Act.KL} measures the divergence between the simulator's client action distribution and the ground-truth action distribution, where lower values indicate better behavioral distribution matching.

\begin{table}[t]
\centering
\caption{Simulator fidelity evaluation on CAMI. Consistency and Realism are rated on a 1--5 scale. Quality is the average of Consistency and Realism. Act.KL measures action distribution divergence from ground truth; lower is better.}
\label{tab:simulator}
\begin{tabular}{lcccc}
\toprule
\textbf{Simulator} & \textbf{Consist.}$\uparrow$ & \textbf{Realism}$\uparrow$ & \textbf{Quality}$\uparrow$ & \textbf{Act.KL}$\downarrow$ \\
\midrule
Profile-based & 3.58 & 1.82 & 2.70 & --- \\
Pro+Act-based & 3.68 & 2.26 & 2.97 & 0.700 \\
CAMI dynamic & 3.34 & 2.68 & 3.01 & \textbf{0.699} \\
Yang et al. & 3.68 & 3.00 & 3.34 & 3.546 \\
\midrule
\textbf{Ours} & \textbf{4.39} & \textbf{3.76} & \textbf{4.07} & 0.839 \\
\bottomrule
\end{tabular}
\end{table}

\paragraph{Results.}
As shown in Table~\ref{tab:simulator}, DynPatient achieves the highest Consistency, Realism, and overall Quality among all evaluated simulators.
Compared with Yang et al., DynPatient improves Consistency from 3.68 to 4.39, Realism from 3.00 to 3.76, and Quality from 3.34 to 4.07.
Its Act.KL is also substantially lower than Yang et al. (0.839 vs. 3.546), indicating a closer match to the real client action distribution.
Although CAMI dynamic and Pro+Act-based obtain slightly lower Act.KL, their response quality scores are much lower.
These results suggest that DynPatient provides a stronger balance between behavioral distribution matching and naturalistic response quality, supporting its use as the shared dynamic evaluation environment.

\paragraph{Limitations.}
Despite these validation results, DynPatient remains an approximation of real client behavior.
Therefore, we use the simulator primarily for relative comparison between counselor policies under a controlled environment.
We do not interpret simulator-based results as direct evidence of real-world clinical effectiveness.

\section{Case Study}
\label{app:case_study}
\begin{figure}[h]
    \centering
    \includegraphics[width=1\linewidth]{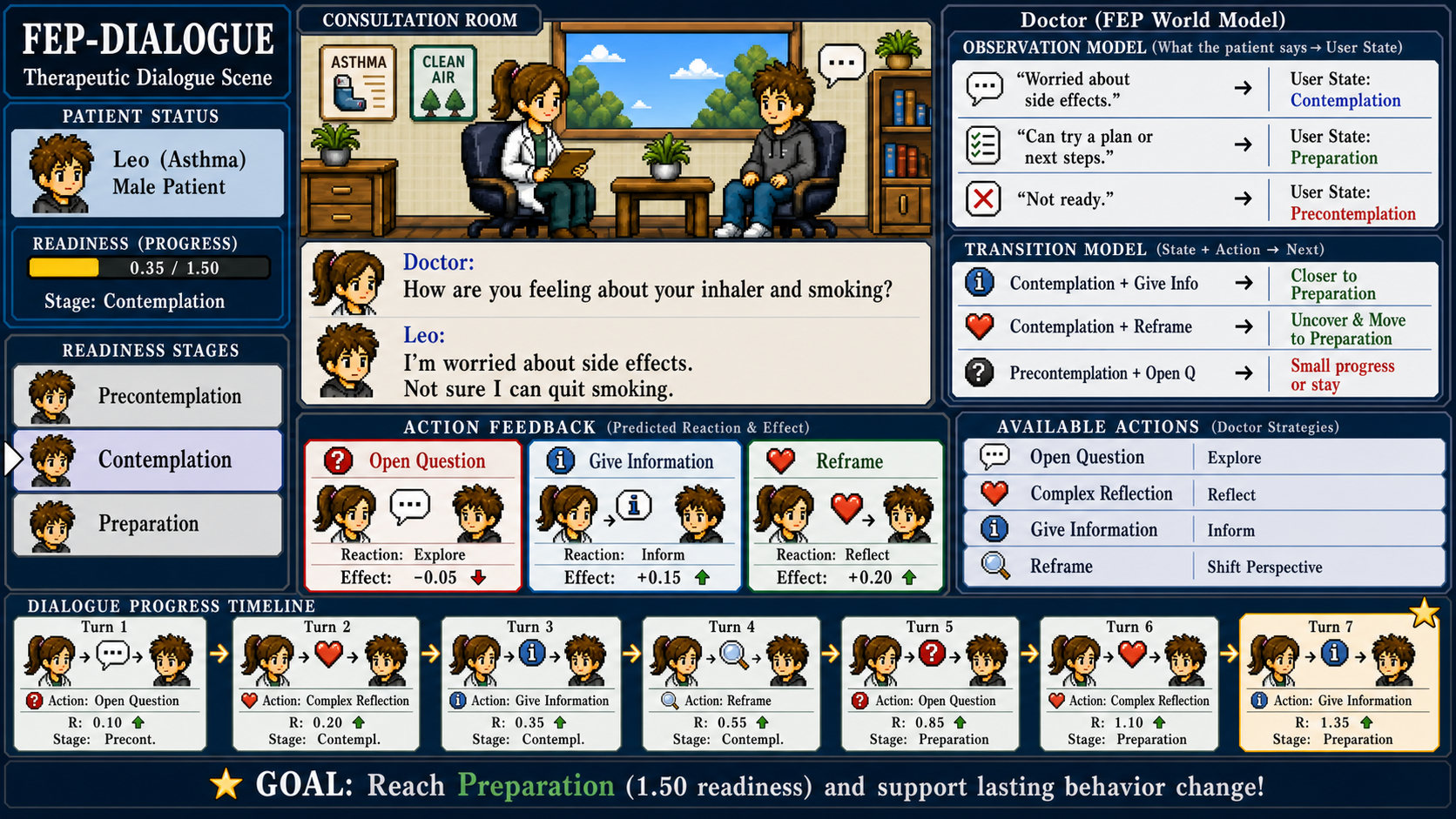}
    \caption{A simulated MI-coaching session between a counselor and a patient}
    \label{fig:your_figure}
\end{figure}

Figure~\ref{fig:your_figure} and Table~\ref{tab:app_case_study} provide the full qualitative case study discussed in Section~\ref{sec:analysis}.
The example compares Base Counselor, DIIR, and PUMA on the same simulated client.
Base Counselor and DIIR both fall into repetitive questioning, while PUMA first explores the user's latent concerns and then selects targeted actions that produce a sharp readiness increase.

\begin{table*}[t]
\centering
\small
\caption{Case study of dynamic action selection. Base Counselor and DIIR fall into repetitive questioning, while PUMA first explores the user's latent concerns and then selects targeted actions that trigger a sharp readiness increase.}
\label{tab:app_case_study}
\begin{tabular}{p{0.10\linewidth}p{0.18\linewidth}p{0.28\linewidth}p{0.16\linewidth}p{0.20\linewidth}}
\toprule
\textbf{Method / Turn} 
& \textbf{Selected Action} 
& \textbf{Representative Interaction} 
& \textbf{Readiness / Stage} 
& \textbf{Mechanism} \\
\midrule

Base, $t_{1}$--$t_{20}$
& Mostly Open Question
& The counselor repeatedly asks variants of how the user feels about the long-term impact.
The user keeps giving similar contemplative responses.
& $0.01 \rightarrow 0.44$; stuck in contemplation
& The dialogue lacks action diversity and fails to elicit new evidence.
Only 3/6 triggers are discovered. \\

DIIR, $t_{1}$--$t_{20}$
& Open / Closed Question
& DIIR starts with varied questions, but from later turns repeatedly asks how the user feels about the inhaler decision.
& $0.01 \rightarrow 0.47$; stuck in contemplation
& Retrieved exemplars lead to a repetitive loop.
The system discovers the same 3/6 triggers as Base. \\

\midrule

PUMA, $t_{1}$--$t_{3}$
& Open Question + Complex Reflection
& The counselor asks what would make the user feel more ready and reflects the user's worry about Sarah's health, inhaler side effects, and making the right decision.
& $0.01 \rightarrow -0.05$; precontemplation
& PUMA temporarily prioritizes epistemic exploration over immediate task advancement, using early turns to infer the user's latent concerns. \\

PUMA, $t_{4}$--$t_{5}$
& Open Question + Give Information
& The counselor asks what specific information would make the user feel more in control, then explains that chronic inhalers reduce inflammation and are generally safe when used as prescribed.
& $0.00 \rightarrow 0.06$; contemplation
& The refined belief indicates that factual uncertainty is blocking progress, so PUMA selects information provision instead of another generic question. \\

PUMA, $t_{6}$--$t_{7}$
& Reframe + Open Question
& The counselor reframes the issue toward reducing secondhand smoke exposure and asks about concrete strategies.
The user mentions smoking only when Sarah is at school and changing clothes before she comes home.
& $0.06 \rightarrow 1.43$; contemplation
& Reframing shifts the dialogue from inhaler fear to actionable asthma-management strategies, unlocking high-value triggers and producing a readiness spike. \\

PUMA, $t_{8}$
& Complex Reflection
& The counselor reflects the user's concrete efforts.
The user asks for regular check-ins and resources for quitting smoking.
& $1.69$; preparation
& PUMA consolidates the user's emerging plan and moves the dialogue into an action-ready state. \\

\bottomrule
\end{tabular}
\end{table*}

\section{Cross-Dataset Generalizability Study}
\label{sec:generalizability}

This appendix provides the full results for \textbf{RQ4}, examining whether the proposed user-state modeling framework can generalize beyond the main CAMI setting.
While the main experiments evaluate the full PUMA framework in a simulator-based dynamic environment, this study isolates the user-state modeling component and tests whether its current-state estimation and next-state prediction abilities transfer to a different dataset and user-state annotation scheme.
To this end, we conduct a cross-dataset experiment on AnnoMI, a real motivational interviewing dataset with utterance-level client-talk annotations.

\paragraph{Dataset.}
We evaluate on \textbf{AnnoMI}~\citep{wu2022anno}, which contains 133 real motivational interviewing counseling transcripts.
The dataset provides expert annotations for client talk type, including \emph{neutral}, \emph{change}, and \emph{sustain}.
Different from CAMI, which uses Transtheoretical-Model stages such as \emph{precontemplation}, \emph{contemplation}, and \emph{preparation}, AnnoMI requires the model to infer the client's motivational orientation from local dialogue signals.
We use the first 50\% of each session as warmup context and the remaining 50\% for evaluation, resulting in 2,382 evaluation turns.

\paragraph{Evaluation protocol.}
We evaluate two state-oriented metrics.
\textbf{Curr-Acc} measures whether the model correctly infers the current client-talk state.
\textbf{Next-Acc} measures whether the model correctly predicts the next client-talk state.
This setting is not intended to reproduce the full dynamic simulation used in the main CAMI experiments.
Instead, it provides a controlled test of whether the proposed user-state belief update and prediction mechanism can transfer to a new state space.
The core user-state modeling procedure remains unchanged, while only the dataset-specific state labels are adapted from CAMI stages to AnnoMI client-talk types.

\paragraph{Compared methods.}
We compare PUMA with long-prompt baselines under two backbone scales, Qwen3-8B and Qwen3-32B.
The long-prompt baseline directly conditions the model on the warmup context and dialogue history, without explicitly maintaining a latent belief state or using the FEP-based inference objective.
PUMA uses Qwen3-8B as the backbone, enabling us to test whether the proposed inference mechanism can outperform a larger long-prompt model.
The detailed prompts for the state-inference baselines are provided in Appendix~\ref{app:prompts_baselines_state}.

\section{Prompt Templates}
\label{app:prompts}
\subsection{PUMA Method Prompts: State Inference, World Model, Action, Generator and State Prediction}
\label{app:prompts_fep}

This appendix provides the main prompt templates used by the PUMA method.

\subsubsection{State Inference}
Due to space limitations, we report abridged prompt templates in the appendix.
The templates preserve the task instruction, label definitions, inference logic,
input interface, and output constraints, while omitting instance-specific input
values and generated outputs.
\label{app:prompts_qfilter}
\begin{promptbox}{State Inference: System Prompt }
You are a core engine responsible for executing
**Incremental Bayesian Belief Updating** over a dual-model system:
- **Observation Model**: p(o|s)
- **Transition Model**: p(s'|s,a)

Your primary objective is to minimize **Variational Free Energy (Delta F)**
under a **strict steady-state constraint**. Large-scale structural overhauls
that destroy existing structures are strictly forbidden; only local, minimized
iterative corrections are allowed.

Return ONLY one valid JSON object.
Do NOT output `<think>`, `</think>`, markdown fences, comments, headings, or
any prose before or after the JSON.
If uncertain, emit the smallest valid JSON matching the schema with empty
arrays where needed.
\end{promptbox}

\begin{promptbox}{State Inference: User Prompt}
## Core Data Model (q_t)
**q_t** is a set of rows {"abstract_cue": "state"} where:
* cue: highly abstract identifier (snake_case) summarizing observed
  regularities before/after the update; NO semantic content or quotes.
* state: must be in {precontemplation, contemplation, preparation}.
* regularity_observation: structure-only contrast of o_t against the prior
  state (pace, length, pauses); NO quotes or paraphrasing.

## Inference Logic (Surprise Minimization Framework)
1. Expectation Generation: from q(s_{t-1}), a_{t-1}, transition prior p,
   predict the structural regularities of the next move.
2. Observation & Surprise Calculation: assimilate o_t; compute the structural
   prediction error vs Step 1 expectations as non-semantic regularities.
3. State Update: update q to absorb the surprise; q_delta arrays may contain
   MULTIPLE add/modify/drop entries simultaneously.

## Strict Rules
- Single valid JSON; no markdown fences; escape internal " as \".
- cue tokens are short snake_case identifiers.
- inference_entropy in [0,1].

## State Boundary Rules
- precontemplation: denies/minimizes the problem, externalizes blame, refuses
  to engage, or shows no awareness of personal risk.
- contemplation: acknowledges some risk/problem/consequence, even partially,
  but has no concrete next step.
- preparation: expresses a concrete action, plan, quota, frequency target,
  time horizon, help-seeking step, or replacement strategy.

Disambiguation:
1. Acknowledgment, even reluctant, -> contemplation, not precontemplation.
2. Tentative planning ("could", "should", "I'll try") -> preparation.
3. Brief responses ("yeah", "I don't know") inherit context; do not regress.

## Update Operations (q_delta)
Each operation is factored into:
| Component             | Definition                         | Examples                              |
| transition_semantic   | abstract latent transition type    | resistance_to_engagement              |
| behavior_pattern      | de-semanticized observable pattern | short_response_with_hedging           |
| state_update          | inferred state + confidence        | {"state":"neutral","confidence":0.85} |

ADD creates a new cue; MODIFY anchors to an existing cue_keyword and updates
its latent interpretation; DROP marks a prior regularity as
contradicted | decayed | overridden.

## Input Interface
The full prompt is instantiated with the following fields:
\[
q(s_{t-1}),\ a_{t-1},\ p(s_t\mid s_{t-1},a_{t-1}),\ p(o\mid s),
\text{ short-term memory},\ o_t.
\]
Instance-specific field values are omitted for space.

## Output Interface
The model outputs a valid JSON object containing:
\texttt{client\_talk\_type},
\texttt{inference\_entropy},
\texttt{q\_abstract},
\texttt{regularity\_observation}, and
\texttt{q\_delta}.
The \texttt{q\_delta} field contains three update operations:
\texttt{add}, \texttt{modify}, and \texttt{drop}.

\end{promptbox}

\subsubsection{World Model Update}
\label{app:prompts_pfilter}
Due to space limitations, we report abridged prompt templates in the appendix.
The templates preserve the task instruction, update logic, input interface,
and output constraints, while omitting instance-specific input values and
generated outputs.
\begin{promptbox}{World Model Update: System Prompt}
You are a core engine responsible for executing
**Incremental Bayesian Belief Updating** over a dual-model system:
- **Observation Model**: p(o|s)
- **Transition Model**: p(s'|s,a)

Your primary objective is to minimize **Variational Free Energy (Delta F)**
under a **strict steady-state constraint**. Large-scale structural overhauls
that destroy existing structures are strictly forbidden; only local, minimized
iterative corrections are allowed.

Return ONLY one valid JSON object.
Do NOT output `<think>`, `</think>`, markdown fences, comments, headings, or
any prose before or after the JSON.
If uncertain, emit the smallest valid JSON matching the schema with empty
arrays where needed.
\end{promptbox}

\begin{promptbox}{World Model Update: User Prompt}
## 1. Free Energy Decomposition (Expectation -> Contrast -> Update)

### Phase 1: Prior Expectation Generation
- Behavioral Expectation: given q(s_t) and current p(o|s), predict reply shape.
- Transition Expectation: given (s_{t-1}, a_{t-1}) and p(s'|s,a), predict s_t.

### Phase 2: Error Contrast & Calculation
- Observation Prediction Error: -log p(o_t | E[q(s_t)]).
- Transition Surprisal: D_KL( q(s_t) || p(s_t | s_{t-1}, a_{t-1}) ).

### Phase 3: Error Minimization Update
- Small DeltaF -> retain parameters.
- Large DeltaF -> minimized corrections via add/modify/drop.

## 2. Dual-Model Maintenance Spec
Each rule is uniquely identified by a snake_case `rule_index`.

### 2.1 Observation model p(o|s) -- {index -> state -> behavior}
- reply_shape MUST be a fully de-semanticized behavioral pattern
  (e.g. short_ack, deflection); no verbatim quotes.

### 2.2 Transition model p(s'|s,a) -- {index -> (s_from, action_a, s_to)}
- action_a MUST come from the provided action vocabulary.

## Input Interface
The full prompt is instantiated with the following fields:
\[
\text{long-term memory},\ \text{recent history},\ p(o\mid s),\
p(s'\mid s,a),\ q(s_{t-1}),\ a_{t-1},\ q(s_t),\ o_t,
\text{ current system utterance},\ \mathcal{A}.
\]
Instance-specific field values are omitted for space.

## Output Interface
The model outputs a valid JSON object containing:
\texttt{therapist\_action},
\texttt{transition},
\texttt{fep\_error\_calculation},
\texttt{observation\_model\_delta}, and
\texttt{transition\_model\_delta}.

The \texttt{fep\_error\_calculation} field records the contrast between
expected and observed behavior, as well as the contrast between the expected
and inferred user state.

The two delta fields specify updates to the observation model and transition
model, respectively. Each delta supports three operations:
\texttt{add}, \texttt{modify}, and \texttt{drop}.
\end{promptbox}

\subsubsection{EFE Action Selection}
\label{app:prompts_efe}

\begin{promptbox}{EFE Action Selection: System Prompt}
You are an expert MI strategy selector. Return only the action name.
\end{promptbox}

\begin{promptbox}{EFE Action Selection: User Prompt}
You are an MI counselor selecting your next therapeutic action.

Current patient state: {current_stage}

Patient observation model (what you've learned about this patient):
{obs_str[:800] or "(no observations yet)"}

Transition model (what you've learned about action effects):
{tr_str[:800] or "(no transition data yet)"}

{diversity_note}

Available actions:
{actions_list}    # MISC 17-class vocabulary

Select the SINGLE best action for this moment. Consider:
1. What do you still need to learn about this patient? (exploration)
2. What action would best advance the patient toward change? (exploitation)
3. Avoid repeating the same action -- try different approaches.

Return ONLY the action name, nothing else.
\end{promptbox}

\subsubsection{Counselor Response Generator}
\label{app:prompts_generator}

\begin{promptbox}{Counselor Response Generator: System Prompt}
As a communication expert with outstanding communication habits, you embody
the role of {agent_name} throughout the following dialogues. Here are some of
your distinctive personal traits: {agent_traits}.
\end{promptbox}

\begin{promptbox}{Counselor Response Generator: User Prompt}
<CONTEXT>
Drawing from your recent conversation with {usr_name}:
{context}

<MEMORY>
The memories linked to the ongoing conversation are:
{memories}

<USER_TRAITS>
During the conversation process between you and {usr_name} in the past, you
found that the {usr_name} has the following characteristics:
{user_traits}

Now, please role-play as {agent_name} to continue the dialogue between
{agent_name} and {usr_name}.

{usr_name} just said: {inquiry}

MI strategy to use: [{therapist_action}] -- apply this counseling approach.

Please respond to {usr_name}'s statement using the following format
(maximum 30 words, must be in English):

RESPONSE:
\end{promptbox}

\subsubsection{State Prediction}
\label{app:prompts_planner}

\begin{promptbox}{State Prediction: System Prompt}
You are a transition model for patient latent state in motivational-
interviewing dialogue (AnnoMI-style). Latent s is summarized by
client_talk_type in {precontemplation, contemplation, preparation}.

Given the current belief, observation/transition history, a discrete therapist
action label from the run vocabulary, and the therapist's actual utterance,
predict client_talk_type *after* that action and how the patient may surface
behaviorally next.

Reply with nothing before the JSON: output exactly one JSON object on the
first line, then stop. No markdown fences, no commentary, no thinking tags.
\end{promptbox}

\begin{promptbox}{State Prediction: User Prompt}
Current patient state:
"""{current_state}"""

Current observation model p(o|s):
"""{observation_model}"""

Current transition model p(s'|s,a):
"""{transition_model}"""

Therapist action label a (must be one of: {actions_line}):
{action}

Therapist's actual message (grounding for a):
"""{therapist_utterance}"""

Output JSON with exactly these keys:
{
  "latent_mi_next": { {inner_latent_mi} },
  "expected_behavior_cues": "how the patient may speak or act next (short)",
  "rationale": "one short sentence linking action a to the predicted shift",
  "confidence": 0.0-1.0
}

Rules:
- latent_mi_next must contain exactly: {dims_line};
  client_talk_type must be one of: {state_values_enum}
- Be conservative: small steps from the current belief unless the utterance
  strongly implies a jump.
- For CAMI stages: tentative planning ("could", "will try", "no more than X")
  is sufficient for `preparation`. Do not regress preparation -> contemplation
  unless the therapist plausibly undermines commitment. Do not regress
  contemplation -> precontemplation unless the client actively re-denies the
  problem.
\end{promptbox}
The planner output for the current turn is stored as the prior used in the
\textbf{next} turn's soft fusion:
\[
p_{\text{fused}}(s_t)= (1-\beta)\,p_{\text{obs}}(s_t)+\beta\,p_{\text{prior}}(s_t),
\]
with $\beta=0.35$ and an utterance-length-aware width on $p_{\text{obs}}$
($\alpha\in\{0.50,0.65,0.75,0.85\}$ for $n_{\text{words}}<6$, $<12$ or
hedge-bearing, $<25$, otherwise).

\subsection{DynPatient: Profile-Grounded Patient Simulator}
\label{app:prompts_sim}

\subsubsection{Patient Action Selection}
\label{app:prompts_sim_action}

\begin{promptbox}{Patient Action Selection: System Prompt}
You are simulating a real patient's behavioral response in a counseling
session about {topic} ({behavior}).
The patient's background: {personas_str}
/no_think
\end{promptbox}

\begin{promptbox}{Patient Action Selection: User Prompt}
Recent conversation:
{context}     # last 4 turns

The counselor just said: "{counselor_utt}"

The counselor just touched on: {matched_trigger_texts[:2]}   # if any

The patient is in the {stage} stage. The probability distribution of this
patient's typical actions is:
{dist_str}    # Dirichlet-smoothed P(action | stage), per-profile

IMPORTANT: Do NOT always pick the most probable action. Real patients vary
their responses naturally. Consider what the counselor just said and pick the
action that fits THIS specific moment. For example:
- If the counselor asks a factual question -> Inform
- If the counselor reflects back feelings -> Acknowledge or Hesitate
- If the counselor challenges          -> Downplay, Deny, or Doubt
- If the counselor discusses plans     -> Plan or Reject

Use the probability distribution as a rough guide for this patient's
personality, but let the conversation context determine the specific choice.

Which action does the patient take right now? Reply with ONLY the action name.
\end{promptbox}

The 11-class patient action vocabulary is
\{Inform, Engage, Downplay, Blame, Deny, Acknowledge, Hesitate, Doubt, Plan,
Accept, Reject\}.

\subsubsection{Patient Response Generation}
\label{app:prompts_sim_response}

\begin{promptbox}{Patient Response Generation: System Prompt}
You are a patient in a counseling session about {topic} ({behavior}).

Your background:
{personas_str}      # up to 5 persona sentences from the profile

Your beliefs:
{beliefs_str}       # up to 4 belief sentences from the profile

{trigger_context}   # any matched profile triggers (Beliefs/Motivation/Plans)

Rules: Reply in first person, 1-3 sentences, natural conversational tone with
hesitations and fillers like real speech. Directly respond to what the
counselor just said. Do NOT mention therapy techniques or that you're in
counseling.
/no_think
\end{promptbox}

\begin{promptbox}{Patient Response Generation: User Prompt}
Conversation so far:
{context}     # last 6 turns

The counselor just said: "{counselor_utt}"

Your action: {patient_action} -- {instruction}    # action_instructions[a]

Respond as the patient (1-3 sentences):
\end{promptbox}

\begin{promptbox}{Patient Response Generation: Action-Conditioned Instructions}
Inform:      Share factual information about your situation related to what
             was asked.
Engage:      Show you're listening and respond to what the counselor said.
Downplay:    Minimize the issue. Show you don't think it's a big deal.
Blame:       Deflect responsibility to external factors.
Deny:        Refuse to acknowledge there's a problem.
Acknowledge: Show you're starting to see the counselor's point, but remain
             tentative.
Hesitate:    Express uncertainty about change. You want to but aren't sure.
Doubt:       Question whether change is possible or worth it.
Plan:        Discuss concrete steps you might take toward change.
Accept:      Agree to try something. Show readiness to move forward.
Reject:      Decline what's being suggested.
\end{promptbox}

\subsection{MISC 17-Class Counselor-Action Classifier}
\label{app:prompts_misc}

This classifier is invoked once per counselor turn during dynamic evaluation
to map each counselor utterance into the MISC 17-class vocabulary, which is
the action label fed into the FEP world model.

\begin{promptbox}{MISC 17-Class Counselor-Action Classifier: System Prompt}
You are an expert Motivational Interviewing (MI) behavior coder trained on
the MI Skill Code (MISC, Miller et al., 2002).

Your task: classify a single counselor utterance into EXACTLY ONE of the
following 17 MISC behavior categories.

Categories:
- Open Question:           open-ended question that invites elaboration.
- Closed Question:         yes/no or short-answer question.
- Simple Reflection:       restatement; adds nothing.
- Complex Reflection:      adds meaning, infers, summarizes themes.
- Affirm:                  positive comment acknowledging strengths/efforts.
- Reframe:                 suggests a different meaning for an experience.
- Support:                 generally supportive comment.
- Emphasize Control:       emphasizes the client's freedom of choice.
- Advise with Permission:  advice given AFTER permission was sought.
- Facilitate:              short acknowledgment ("Mm-hmm", "Tell me more").
- Give Information:        provides factual information.
- Structure:               session-flow / transitions / logistics.
- Raise Concern:           collaborative concern about the client's plan.
- Confront:                disagrees, argues, criticizes, judges.
- Direct:                  imperative order/command.
- Warn:                    warning of negative consequences.
- Advise without Permission: unsolicited advice.

Important rules:
- Choose the SINGLE most dominant behavior in the utterance.
- If the utterance contains a question AND a reflection, prioritize whichever
  is the main intent.
- "Advise with Permission" requires evidence that permission was sought.
- "Facilitate" is for very short acknowledgments only.
- Return ONLY valid JSON: {"label": "<one of the 17 categories>"}

/no_think
\end{promptbox}

\begin{promptbox}{MISC 17-Class Counselor-Action Classifier: User Prompt}
Dialogue context:
{context[-500:]}

Client state: {state}

Counselor utterance to classify:
"{counselor_utt}"

Return JSON: {"label": "<MISC category>"}
\end{promptbox}

\subsection{Long-Prompt Static State-Inference Baselines}
\label{app:prompts_baselines_state}

\begin{promptbox}{Long-Prompt Baseline: System Prompt}
You are a readiness stage classifier using the Transtheoretical Model.
Infer the patient's current readiness stage and likely immediate next stage
after the therapist's reply.

Use ONLY these labels: precontemplation, contemplation, preparation.
- precontemplation: the patient does not recognize the behavior as
  problematic and tends to defend the status quo.
- contemplation:    the patient recognizes some problem signals but remains
  ambivalent and mixed about change.
- preparation:      the patient is ready to take action and expresses
  concrete movement toward change.

Return strict JSON.
\end{promptbox}

\begin{promptbox}{Long-Prompt Baseline: User Prompt}
Given the full interaction history, infer:
1. `current_state`: the label of the CURRENT patient utterance.
2. `next_state`:    the label the patient is MOST likely to express in their
   NEXT utterance after the doctor's latest reply.

Output JSON exactly as:
{"current_state":"precontemplation|contemplation|preparation",
 "next_state":   "precontemplation|contemplation|preparation",
 "rationale":   "<brief>"}

Full history:
{history_text}
\end{promptbox}

\end{document}